\documentclass[11pt]{article}
\usepackage[final]{acl} 

\usepackage{amsmath}
\usepackage{times}
\usepackage{latexsym}
\usepackage[T1]{fontenc}
\usepackage[utf8]{inputenc}
\usepackage{tablefootnote}
\usepackage{microtype}
\usepackage{booktabs}
\usepackage{inconsolata}
\usepackage{graphicx}
\usepackage{amssymb}
\usepackage{pifont}
\usepackage[most]{tcolorbox}

\title{More Agents Improve Math Problem Solving but Adversarial Robustness Gap Persists}

\author{Khashayar Alavi\textsuperscript{$\clubsuit$}\thanks{Equal contribution} \hspace{3mm} \textbf{Zhastay Yeltay}\textsuperscript{$\clubsuit$}\footnotemark[1] \hspace{3mm} \textbf{Lucie Flek}\textsuperscript{$\clubsuit\spadesuit$} \hspace{3mm} \textbf{Akbar Karimi}\textsuperscript{$\clubsuit\spadesuit$} \\ 
    \textsuperscript{$\clubsuit$}Bonn-Aachen International Center for Information Technology,  University of Bonn, Germany \\
    \textsuperscript{$\spadesuit$}Lamarr Institute for Machine Learning and Artificial Intelligence, Germany\\
    \texttt{\{s76kalav, s06zyelt\}@uni-bonn.de}\\
    \texttt{ak@bit.uni-bonn.de}}

\begin{document}
\maketitle
\begin{abstract}
When LLM agents work together, they seem to be more powerful than a single LLM in mathematical question answering. However, are they also more robust to adversarial inputs? We investigate this question using adversarially perturbed math questions. These perturbations include punctuation noise with three intensities ($10\%$, $30\%$, $50\%$), plus real-world and human-like typos (\texttt{WikiTypo}, \texttt{R\textsuperscript{2}ATA)}. Using a unified sampling-and-voting framework (Agent Forest), we evaluate six open-source models (Qwen3-4B/14B, Llama3.1-8B, Mistral-7B, Gemma3-4B/12B) across four benchmarks (GSM8K, MATH, MMLU–Math, MultiArith), with various numbers of agents $n\in\{1,2,5,10,15,20,25\}$. Our findings show that 1) Noise type matters: punctuation noise harm scales with its severity, and the human typos remain the dominant bottleneck, yielding the largest gaps to \texttt{Clean} accuracy and the highest attack success rate (ASR) even with a large number of agents; 2) Collaboration reliably improves accuracy as the number of agents, $n$, increases, with the largest gains from $n{=}1$ to $n{=}5$ and diminishing returns beyond $n{\approx}10$. However, the adversarial robustness gap persists regardless of the agent count.
\end{abstract}

\section{Introduction}

\begin{figure}[t]
      \centering
      \includegraphics[width=\linewidth]{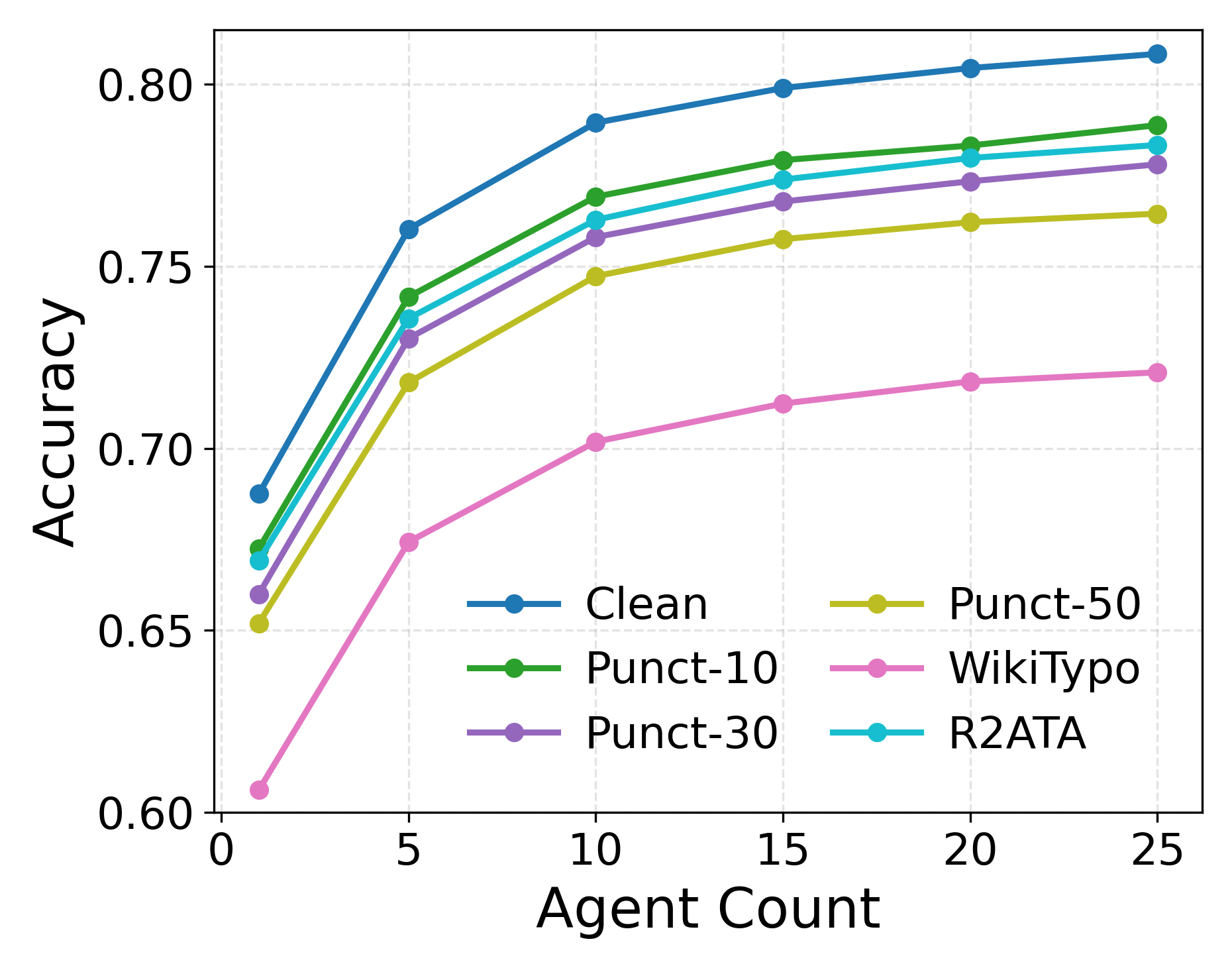}
      \caption{Average accuracy across datasets and models as a function of agent count. Lines denote noise types (\texttt{Clean}, \texttt{Punct-}$10$\texttt{/}$30$\texttt{/}$50$, \texttt{WikiTypo}, \texttt{R\textsuperscript{2}{ATA}}). Markers show mean accuracy across datasets.}
      \label{fig:combined_overall_accuracy.png}
    \end{figure}
    
    Large Language Models (LLMs) have made remarkable progress in solving a variety of reasoning-intensive tasks \cite{zhao2025surveylargelanguagemodels}, including mathematical problem solving. Beyond simply scaling model size, recent work shows that treating multiple LLM instances as \textit{agents} and combining their outputs through debate, voting, or self-consistency can substantially boost performance. \citet{li2024agentsneed} demonstrate that even a simple sampling-and-voting strategy, essentially allowing multiple agents to propose answers and then aggregating them, can rival more sophisticated collaborative frameworks. These findings suggest that collaboration between agents is a powerful and relatively simple way to enhance reasoning accuracy.
    At the same time, another body of research highlights a major weakness of LLMs: their fragility when inputs are perturbed. While the semantic meaning of a math problem may remain unchanged, minor surface-level changes, such as synonym replacements \cite{shi2023largelanguagemodelseasily}, typos \cite{gemmateam2024gemma2improvingopen}, or even random punctuation insertions, can significantly reduce model accuracy. For instance, \citet{abedin2025arithmattack} show that inserting punctuation noise, despite leaving the problem content intact, can cause accuracy drops on well-established benchmarks like GSM8K \cite{cobbe2021trainingverifierssolvemath} and MultiArith \cite{roy2016solvinggeneralarithmeticword}. 
    These two perspectives converge on an important open question: \textit{If multiple agents can enhance reasoning performance on \texttt{Clean} inputs, can they also improve robustness under noisy conditions?} 
    Understanding whether ensemble-based multi-agent setups show resilience to noise would provide valuable insights for designing more reliable and trustworthy LLM systems.
    
    To address this gap, we investigate the robustness of Agent Forest (a multi-LLM-agent setup) to adversarially perturbed math questions. We develop a taxonomy of noise types for adversarial modification and systematically introduce them across multiple reasoning datasets for mathematical question answering, enabling a controlled evaluation of their impact. 
    We use ArithmAttack \cite{abedin2025arithmattack} to generate synthetic noise, while for real-world noise we rely on natural typos from Wikipedia \cite{aliakbarzadeh2025exploringrobustnessmultilingualllms} and adversarial perturbations introduced by the Adversarial Typo Attack \cite{gan2024reasoningrobustnessllmsadversarial}.
    We assess the impact of input noise perturbations on open-source LLMs, specifically Qwen \cite{yang2025qwen3technicalreport}, Llama \cite{llamatechnicalreport}, Mistral \cite{jiang2023mistral7b}, and Gemma \cite{gemmateam2025gemma3technicalreport}, across multiple parameter scales. By doing so, we evaluate not only how noise affects individual models but also whether multi-agent collaboration enhances resilience and preserves reasoning accuracy in the presence of such perturbations.

    Figure \ref{fig:combined_overall_accuracy.png} gives an overview of the results of our investigation. Across all experiments, we observe a consistent, monotonic gain from increasing the number of collaborating agents. Averaged over datasets and models, accuracy rises from $0.6579$ with a single agent to $0.7740$ with $25$ agents, with the largest jump occurring between $one$ and $five$ agents and clear diminishing returns beyond $10$. 
    
    The relative ordering of noise conditions is stable: With the \texttt{Clean} performance at the top, \texttt{Punct-}$10$ (i.e., inserting punctuation noise in as many spaces as $10$ percent of the question length) causes a consistent accuracy drop, then even larger drops are caused by \texttt{Punct-}$30$ and \texttt{Punct-}$50$, while \texttt{WikiTypo} is consistently the most challenging, causing a drop of $8\%$ across all the settings with different numbers of agents. \texttt{R\textsuperscript{2}ATA} causes a slightly larger drop than \texttt{Punct-}$10$ in our aggregate.

\section{Related Work}
    Recent work has shown that current open-source large language models (LLMs) exhibit limited resilience to naturally occurring input noise, such as spelling mistakes, typographical errors, and minor punctuation artifacts, raising serious concerns about their reliability in real-world deployments \cite{dong2023revisitinputperturbationproblems, abedin2025arithmattack}.
    A substantial body of research has analyzed the impact of input perturbations across multiple granularities. At the character level, studies examine misspellings and typos \cite{gao2018black, li2018textbugger}; at the word level, attacks involve synonym substitution or word shuffling \cite{garg2020bae, moradi2021evaluating}; and at the sentence level, robustness is tested via paraphrasing or insertion of irrelevant context \cite{shi2023robustness, lanham2023adversarial, arora2025exploring}. Even seemingly innocuous changes, such as random punctuation insertion, have been shown to significantly degrade performance \cite{abedin2025arithmattack}.
    These perturbations have been evaluated across a range of tasks and settings. Prior analyses of LLMs under noisy inputs include text classification \cite{eger2020hero}, natural language inference \cite{formento2021special, formento2023using}, and multilingual scenarios \cite{aliakbarzadeh2025exploringrobustnessmultilingualllms}, offering valuable insights into task-specific vulnerabilities.

    In parallel, a growing line of work \cite{talebirad2023multiagentcollaborationharnessingpower, seo2025spioensembleselectivestrategies, wang2025tdagmultiagentframeworkbased} demonstrates that multi-agent ensembles can substantially boost reasoning performance on \texttt{Clean} inputs. These systems often form complex networks of LLMs, coordinating via peer-to-peer or distributed configurations \cite{tran2025survey}. For example, multi-agent debates allow models to iteratively propose and critique answers \cite{du2023improving}, a process that can be further optimized using sparse communication networks to reduce error propagation \cite{li2024improving}. Despite the sophistication of such topologies, \citet{lu2024blendingneedcheaperbetter} show that simple strategies such as majority voting often rival complex debate frameworks, allowing ensembles of smaller models to match much larger single models. However, the robustness of these collaborative setups remains an open question. \citet{helmi2025modelingresponseconsistencymultiagent} provide preliminary evidence that multi-agent reasoning degrades under high noise, highlighting a critical vulnerability.
    
    Together, these observations reveal a severe gap. While both noise robustness and multi-agent collaboration are active research areas, their intersection has not been systematically explored. Although recent frameworks continue to introduce highly complex, multi-round network topologies \cite{becker2025mallm, asad2025reddebate}, understanding their fundamental vulnerabilities first requires a solid, isolated baseline. By focusing on a sampling-and-voting topology, our work directly addresses this gap, establishing a baseline for evaluating the adversarial robustness of multi-LLM-agent systems. 

\section{Methodology}
\begin{figure}[t]
      \centering
      \includegraphics[width=\linewidth]{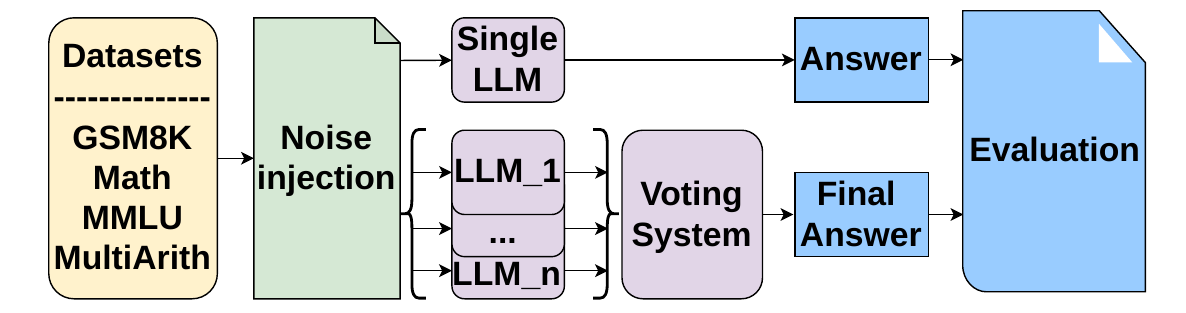}
      \caption{Overview of the experimental framework for evaluating multi-LLM-agent robustness. The system processes input questions from four datasets under  \texttt{Clean} and noisy conditions. For each condition, a single LLM generates multiple independent answers (LLM$_1$ to LLM$_n$), which are then aggregated by majority voting to produce the final answer.}
      \label{fig:test_archi.png}
    \end{figure}

Figure \ref{fig:test_archi.png} depicts our pipeline for evaluating the adversarial robustness of LLMs from noisy dataset generation to answer extraction. 

\subsection{Noisy Dataset Creation}
    To systematically analyze the influence of noise on multi-agent LLMs, we introduce a taxonomy that distinguishes between synthetic noise and human-like noise. This taxonomy enables a structured evaluation of model performance under both artificially generated perturbations and noise patterns that more closely resemble human input.
    \paragraph{Synthetic noise.}Following \texttt{ArithmAttack} \cite{abedin2025arithmattack}, we implement synthetic noise with \textit{punctuation insertion}, where symbols from the set \{".", ",", "!", "?", ";", ":"\} are randomly added to sentences. To study the effect of perturbation strength, each noise type is applied at three intensity levels ($10\%$, $30\%$, and $50\%$), allowing a controlled analysis of model performance under progressively noisier conditions.

    \paragraph{Human-like noise.} To mimic realistic human errors, we consider two complementary approaches. The first is the Adversarial Typo Attack (ATA) \cite{gan2024reasoningrobustnessllmsadversarial}, which evaluates the robustness of LLMs to typographical errors by iteratively introducing common character-level mistakes typically caused by keyboard usage. These include letter duplication, character substitution, deletion, or insertion, and removal of spaces, all targeted toward words that are important for the query, with the goal of degrading model performance. For this, we rely on the \texttt{R\textsuperscript{2}ATA} benchmark, which applies ATA to GSM8K, BBH, and MMLU tasks. The second approach leverages naturally occurring spelling mistakes extracted from Wikipedia edit histories \cite{aliakbarzadeh2025exploringrobustnessmultilingualllms}. Specifically, we use the \texttt{WikiTypo} dictionary, which provides a collection of real-world human spelling errors, to replace words in the datasets with authentic typo variants.

\subsection{Agent Forest}
\label{subsec:multi_llm_agent}
    We evaluate multi-agent performance using \emph{Agent Forest} \cite{li2024agentsneed}, which consists of two phases: sampling and voting. 
    For each question $x$ from a dataset, we query the same base model $\mathcal{M}$ multiple times with independent random seeds and fixed decoding settings. Each call acts as one agent and produces one candidate solution $s_i$. After $N$ calls we obtain the set of candidates $S=\{s_1,\dots,s_N\}$.

    \paragraph{Answer extraction and normalization.}
    For arithmetic datasets, we extract the final numeric answer from each $s_i$ using simple patterns that look for common markers such as \texttt{Answer} or \texttt{\textbackslash boxed\{\}}. We then extract the value by removing formatting artifacts, converting textual numbers to numerals, and standardizing whitespace. This yields a canonical answer $\mathrm{canon}(s_i)$ for each sample.

    \paragraph{Aggregation by voting.} To specify the final answer, we aggregate candidates by majority vote over canonical answers.
    
    \paragraph{Protocol and replication.}
    For every model and dataset, we generate exactly $25$ independent single-agent outputs per question with fixed decoding parameters. The threshold for the number of agents was based on a preliminary study, indicating that variance stabilizes well before this point. To study different agent counts $n \in \{1,2,5,10,15,20,25\}$, we partition them into $\lfloor 25/n \rfloor$ disjoint groups of size $n$ in their generation order. Each group forms one independent $n$ agent ensemble whose vote yields one prediction for that question. Thus we obtain $25$ independent predictions for $n=1$, $12$ for $n=2$, $5$ for $n=5$, and $2$ for $n=10$. For $n=15$, $n=20$, and $n=25$, only one ensemble is available, so we use the first $n$ outputs. At the dataset level, we compute accuracy for each independent group and report the mean and the standard deviation across groups when multiple groups exist. No fine-tuning or task-specific supervision is used, and decoding settings are identical for \texttt{Clean} and noisy inputs so that any change in accuracy is attributable to aggregation by Agent Forest.

\subsection{Evaluation Metrics}
    \label{subsec:metrics}
    We report accuracy and Attack Success Rate (ASR). Let $\mathcal{D}=\{(x_i,y_i)\}_{i=1}^{m}$ be a data set with gold answers $y_i$. Let $f_n$ denote the $n$ agent ensemble predictor that maps an input $x$ to a single answer by Agent Forest voting as defined in Section~\ref{subsec:multi_llm_agent}. The metric is:
    
    \paragraph{Accuracy.}
    \texttt{Clean} accuracy for $n$ agents:
    \[
    \mathrm{Acc}(n)
    =\frac{1}{m}\sum_{i=1}^{m}\mathbf{1}\!\left[\,f_n(x_i)=y_i\,\right].
    \]

    \paragraph{Attack Success Rate (ASR).}
    ASR measures the fraction of originally correct predictions that are flipped to incorrect after applying the noise. Formally,
    \[
    \mathrm{ASR}(n)
    =\frac{\sum_{i=1}^{m}\mathbf{1}\!\left[\,f_n(A(x_i)) \neq y_i\,\right]}
    {\sum_{i=1}^{m}\mathbf{1}\!\left[\,f_n(x_i)=y_i\,\right]}.
    \]
    
    ASR is thus defined on the subset of items that the model with $n$ agents solves correctly in the \texttt{Clean} setting and takes values in $[0,1]$. A lower value indicates better robustness.

    \paragraph{Reporting.}
    As described in Section~\ref{subsec:multi_llm_agent}, for each model and dataset, we generate $25$ independent single-agent outputs per question and form disjoint groups to realize different agent counts $n\in\{1,2,5,10,15,20,25\}$. For $n\in\{1,2,5,10\}$ this yields multiple independent groups; we compute $\mathrm{Acc}$ and $\mathrm{ASR}$ for each group and report the mean and the standard deviation across groups. For $n\in\{15,20,25\}$ only one group is available; we report the single value.

\begin{figure*}[t]
      \centering
      \includegraphics[width=\textwidth]{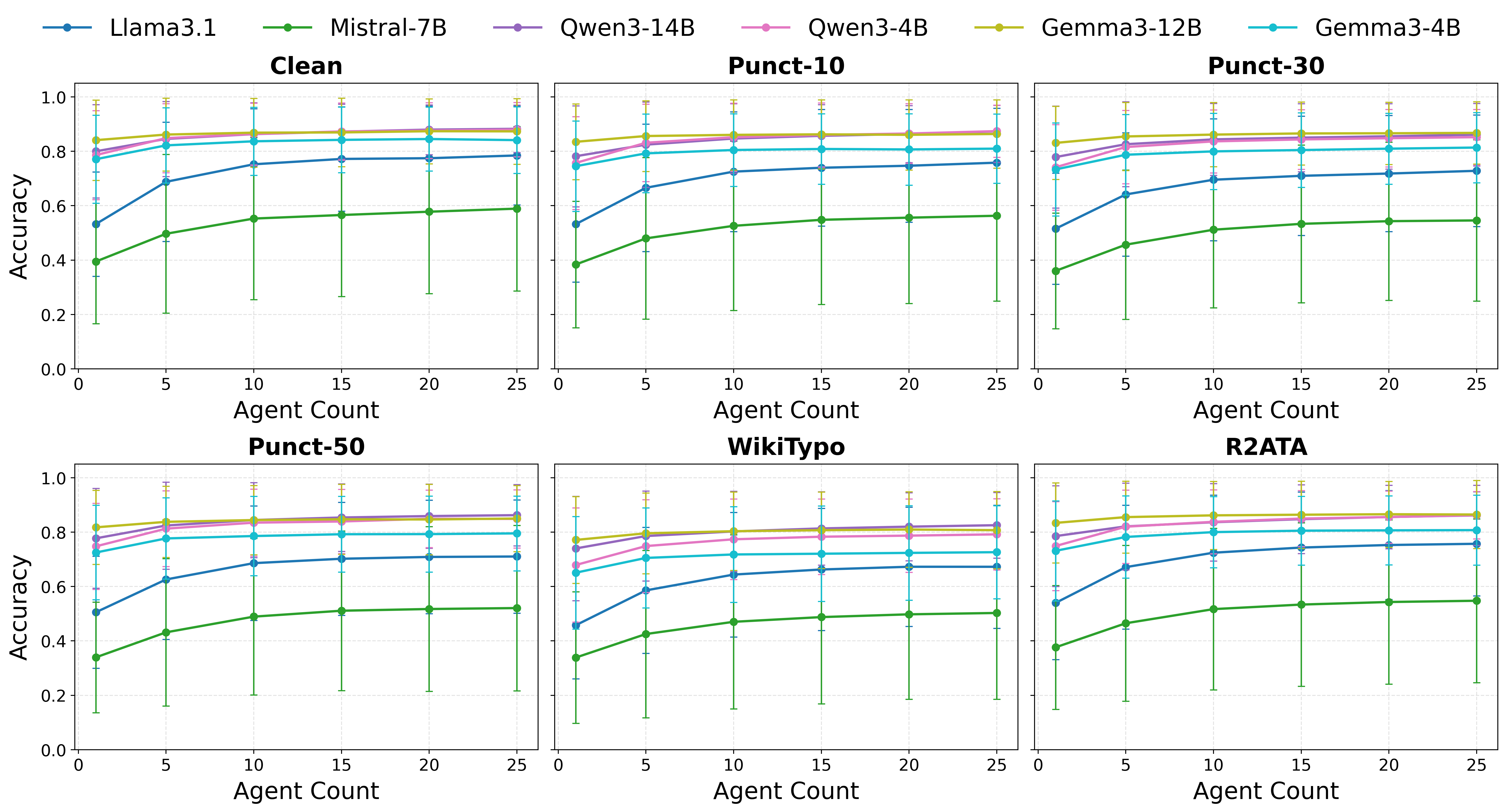}
    \caption{Accuracy averaged across datasets by noise type (\texttt{Clean}, \texttt{Punct-}$10$\texttt{/}$30$\texttt{/}$50$, \texttt{WikiTypo}, \texttt{R\textsuperscript{2}{ATA}}) as a function of agent count. Larger models (Gemma3-12B and Qwen3-14B) maintain higher absolute accuracy across conditions, while smaller (and older) models (Llama3.1 and Mistral-7B) show steeper relative gains with more agents. New small models (Gemma3-4B and Qwen3-4B) show similar performances to their larger counterparts.}
      \label{fig:noise-accuracy-agg}
\end{figure*}

\begin{figure*}[!htb]
      \centering
    \includegraphics[width=\textwidth]{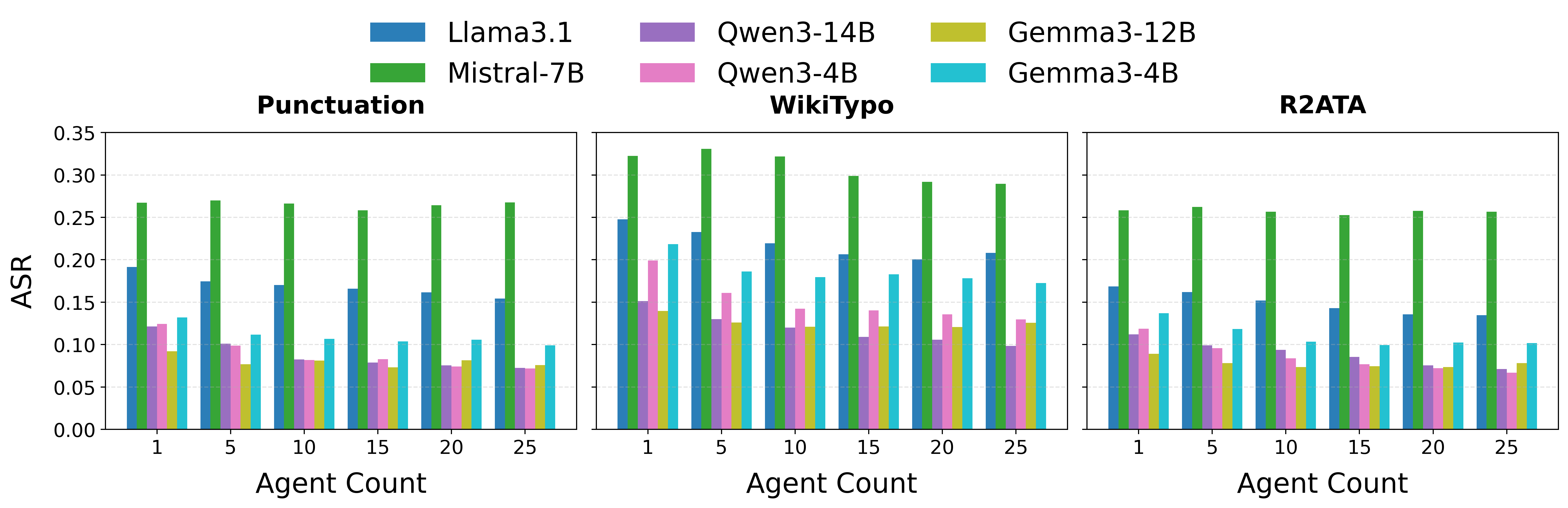}
    \caption{Attack Success Rate averaged across datasets, shown by noise category (\texttt{Punctuation}, \texttt{WikiTypo}, \texttt{R\textsuperscript{2}{ATA}}). Bars represent different models at varying agent counts. Punctuation noise vulnerability is most effectively reduced with more agents. \texttt{WikiTypo} remains the most challenging across all models, and \texttt{R\textsuperscript{2}{ATA}} lies in between, initially harmful for single agents but increasingly mitigated through collaboration.}
      \label{fig:noise-asr-agg}
    \end{figure*}
    
\section{Experiments}
    All experiments were executed on HPC GPU Cluster 8 x NVIDIA GA100 [A100] GPUs with 80GB, running under CUDA 12.2. 
    We used Python~3.10 with the PyTorch~2.2 and HuggingFace Transformers~\cite{wolf2020huggingfacestransformersstateoftheartnatural} libraries. 
    For efficient inference, we employed vLLM \cite{kwon2023efficient}, which allows fast batched decoding for large-scale evaluation. 
    
\subsection{Datasets}
    Evaluation of mathematical reasoning capabilities is conducted across four established benchmarks: \textbf{GSM8K} \cite{cobbe2021trainingverifierssolvemath}, \textbf{MATH} \cite{hendrycks2021measuringmathematicalproblemsolving}, the mathematics subset of \textbf{MMLU} \cite{hendrycks2021measuringmassivemultitasklanguage}, and \textbf{MultiArith} \cite{roy2016solvinggeneralarithmeticword}. This selection provides a comprehensive assessment spanning difficulty levels from basic arithmetic to advanced competition problems while incorporating both open-ended and multiple-choice formats. 
    
    \textbf{GSM8K} \cite{cobbe2021trainingverifierssolvemath} contains 8.5K linguistically diverse grade school math word problems, with evaluation performed on the standard 1.3K test questions requiring multi-step arithmetic reasoning. Its broad adoption as a mathematical problem-solving benchmark offers a standard reference point for studying reasoning robustness under input noise. 
    
    The \textbf{MATH} dataset \cite{hendrycks2021measuringmathematicalproblemsolving} consists of 12.5K challenging mathematics competition problems covering advanced topics, including algebra, geometry, and number theory. These problems require symbolic manipulation and multi-step derivations, enabling assessment of how adversarial noise affects complex problem-solving capabilities compared to elementary-level reasoning.
    
    Evaluation using \textbf{MMLU} \cite{hendrycks2021measuringmassivemultitasklanguage} focuses specifically on the mathematics subsets, which provide multiple-choice questions testing factual recall and mathematical reasoning. This format complements the open-ended generation tasks from other datasets. 
    
    \textbf{MultiArith} \cite{roy2016solvinggeneralarithmeticword} consists of 600 arithmetic word problems requiring multi-step operations, with evaluation on the standard 180-question test set. This dataset serves as a lightweight benchmark for studying noise effects on simple yet compositional reasoning tasks.

\subsection{Models}
    To evaluate a diverse range of capabilities while maintaining computational efficiency, a selection of modern open-source language models of varying parameter sizes was chosen for evaluation to analyze the impact of model scale. The models comprise Qwen3-4B-Instruct and Qwen3-14B-Instruct \cite{yang2025qwen3technicalreport}, Llama-3.1-8B-Instruct \cite{llamatechnicalreport}, Mistral-7B-Instruct-v0.3 \cite{jiang2023mistral7b}, along with Gemma3-4B-IT and Gemma3-12B-IT \cite{gemmateam2025gemma3technicalreport}. These models will be subsequently referred to as \textbf{Qwen3-4B}, \textbf{Qwen3-14B}, \textbf{Llama3.1}, \textbf{Mistral-7B}, \textbf{Gemma3-4B}, and \textbf{Gemma3-12B}, respectively.

\section{Results}
The overall trends, regarding the impact of each adversarial perturbation method, show themselves when we average the results over datasets and models. This can be seen in Figure \ref{fig:combined_overall_accuracy.png}, depicting that all the methods cause a consistent robustness gap, with \texttt{WikiTypo} having the largest one. In the following, we report the results and the trends across the adversarial models, datasets, and adversarial noise, respectively. 

\begin{figure*}[t]
  \centering
  \includegraphics[width=\textwidth]{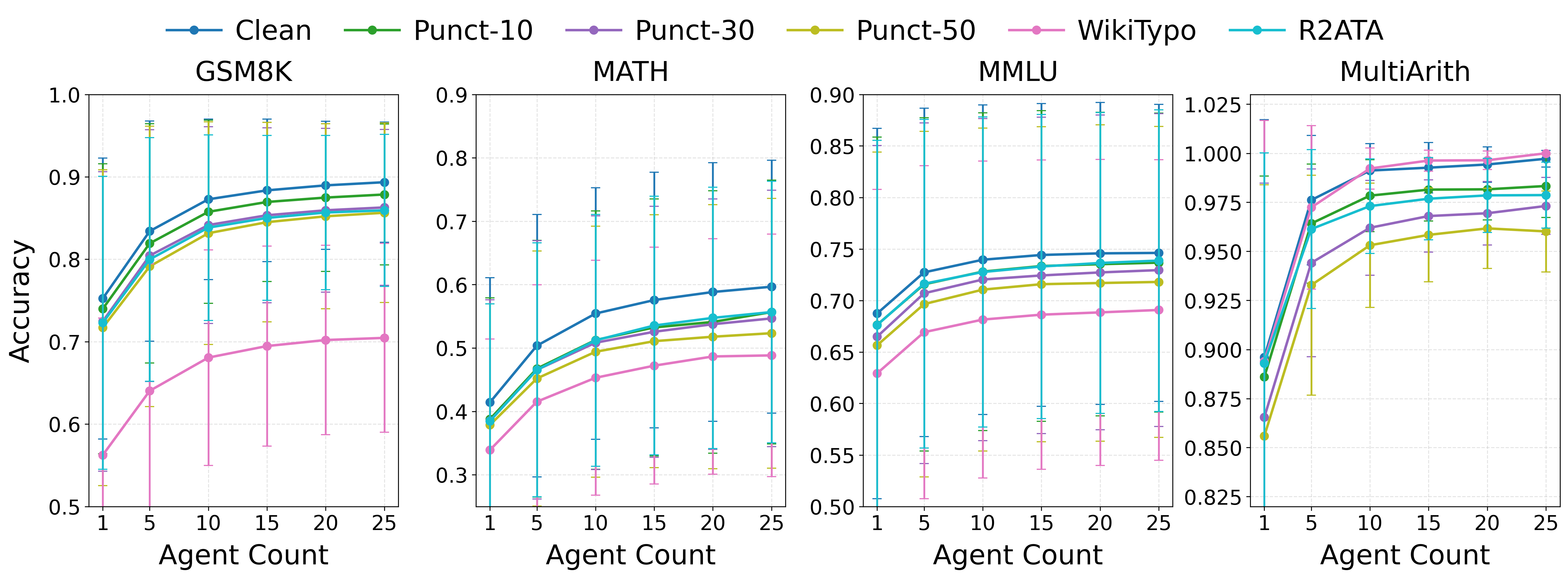}
    \caption{Accuracy averaged across models, shown separately for each dataset as a function of agent count. Collaboration improves accuracy under all noise types, with the largest gains on MATH, MultiArith, and GSM8K. In contrast, MMLU shows smaller absolute drops under noise and saturates more quickly with increasing agent count.}
  \label{fig:dataset-accuracy}
\end{figure*}
\begin{figure*}
  \centering
  \includegraphics[width=\textwidth]{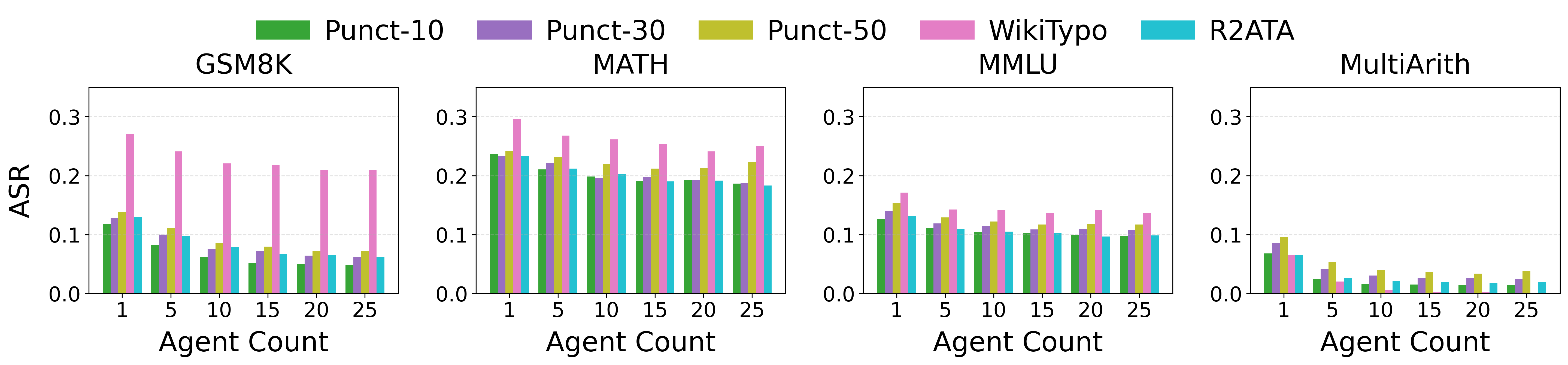}
    \caption{Attack Success Rate averaged across models, shown separately for each dataset. Noise impact differs by task: more difficult datasets, such as MATH and MMLU, exhibit higher ASR, which remains stable with the increase of agents, while easier datasets (GSM8K and MultiArith) show lower ASR values and more benefits from increasing agent count.}
  \label{fig:dataset-asr}
\end{figure*}

\subsection{Models}
    Figure~\ref{fig:noise-accuracy-agg} (also Table \ref{tab:accuracy_results} in detail in Appendix \ref{extended_results}) presents the impact of input perturbations across models and agent counts. 
    Increasing the number of agents reliably improves performance under all noise types. The largest improvements occur between $n{=}1$ and $n{=}5$, with diminishing returns beyond $n{\ge}10$. As the agent count approaches $n{=}15$, the accuracy curves smooth out (we also see this with agent numbers up to 50 as shown in the extended ablation in Figure~\ref{fig:appendix_model_acc_50} in Appendix \ref{extended_results}). The performance delta among $n \in \{25, 30, 40, 50\}$ remains negligible and is comparable to the minor fluctuations observed across runs, indicating that the ensemble effectively reaches a steady state and justifying our focus on agent counts up to $n{=}25$ for the primary evaluation. 

    As shown in Figure \ref{fig:noise-accuracy-agg}, accuracy decreases as punctuation intensity rises from \texttt{Punct-}$10$ to \texttt{Punct-}$50$. This degradation is visible at the single-agent level ($n{=}1$), where models lose between $5\text{--}15$ percentage points compared to the \texttt{Clean} baseline. However, collaboration rapidly compensates for these losses by $n{=}5$, the performance gap between \texttt{Punct-}$10$ and \texttt{Clean} inputs becomes negligible for most model families, and even \texttt{Punct-}$30$ and \texttt{Punct-}$50$ recover a substantial share of their accuracy.

    Despite the accuracy gains from collaboration, noise affects models similarly. As we can see in Figure \ref{fig:noise-asr-agg}, the ASR values stay almost the same when increasing the number of agents. This indicates that under simple surface-level perturbations, the collaboration between agents is as vulnerable to noise as a single LLM.

\begin{figure*}
    \includegraphics[width=1\textwidth]{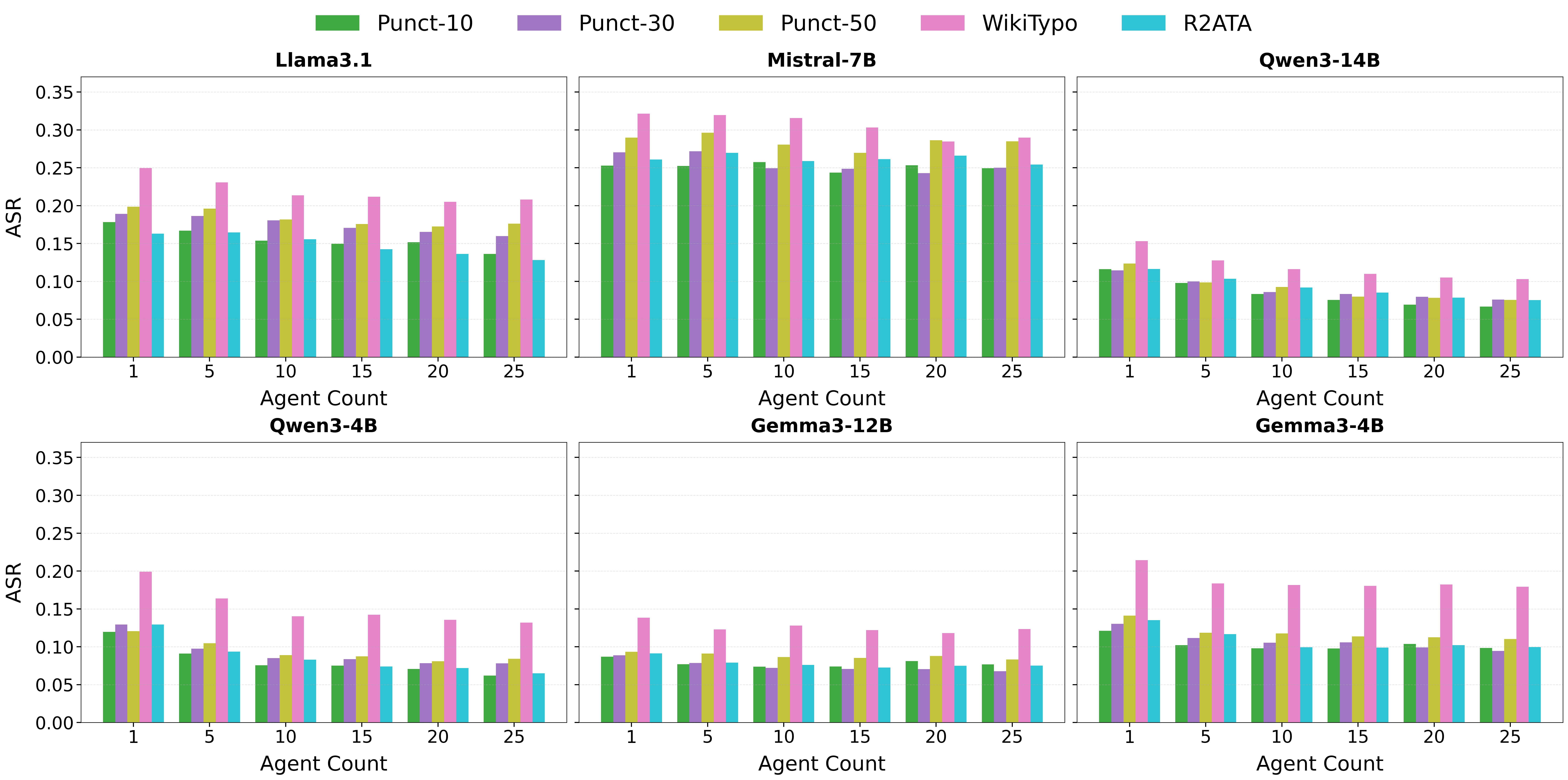}
    \caption{Attack Success Rate averaged across datasets, shown per model family and agent count. Generally, all the investigated noise types consistently decrease the model performance across different numbers of agents. \texttt{WikiTypo}, however, seems to be the strongest across the models and agent counts.}
    \label{fig:combined_datasets_asr_bars_by_model_paper}
\end{figure*}

\subsection{Datasets}
    Figures ~\ref{fig:dataset-accuracy} and ~\ref{fig:dataset-asr} reveal distinct robustness profiles by dataset.
    MATH is the most difficult benchmark in absolute terms. Accuracy starts lowest and improves with more agents, yet gaps to the \texttt{Clean} baseline remain under all types of adversarial inputs even at large agent counts. Its ASR is also the highest among the datasets and hardly declines with increasing $n$, indicating that long, multi-step solutions are sensitive to both synthetic punctuation and human-like typos.
    \\
    For the GSM8K dataset, moving from one to five agents recovers much of the loss from punctuation noise, and ASR drops steadily with $n$. \texttt{WikiTypo} consistently induces the largest degradation at a fixed agent count, suggesting heterogeneous human errors are more disruptive than punctuation insertions on these short, compositional problems.
    \\
    MMLU (math subset) is comparatively stable. Accuracies cluster tightly and rise smoothly with agent count, but its ASR is slightly higher than in GSM8K across noise types.
    \\
    MultiArith is near the ceiling. Accuracy saturates by approximately five agents across all noise types, and ASR is minimal and continues to decrease with $n$. The short, structured arithmetic chains offer limited headroom for noise to reduce performance or for collaboration to yield large additional gains.
    \\
    Overall, easier datasets with simpler tasks (GSM8K, MultiArith) are less affected by noise, which is reflected in their lower ASR values. Any performance degradation can be compensated for more easily through multi-agent collaboration. In contrast, more challenging datasets (MATH and MMLU) suffer greater absolute performance drops under noise, and even with larger agent counts, the recovery remains incomplete, highlighting that task difficulty amplifies the impact of input perturbations.

\subsection{Adversarial noise types}
    Figure~\ref{fig:combined_datasets_asr_bars_by_model_paper} presents the Attack Success Rate (ASR) aggregated across datasets, broken down by model family and agent count. Two major patterns emerge. 
    Across all conditions, we observe an almost stable ordering of impact:  \texttt{WikiTypo} $>$ \texttt{Punct-}$50$ $\ge$ \texttt{Punct-}$30$ $\ge$ \texttt{R\textsuperscript{2}{ATA}} $\ge$ \texttt{Punct-}$10$, with \texttt{R\textsuperscript{2}ATA} typically close to the \texttt{Clean} baseline. This ranking can be more visibly seen in Figure \ref{fig:combined_overall_accuracy.png}. 

However, the punctuation noise types as well as \texttt{R\textsuperscript{2}ATA} seem to generally have a similar negative impact on the performance of all the investigated models. This impact is also somewhat stable as we increase the number of agents.  By contrast, \texttt{WikiTypo} remains the dominant bottleneck. Even at $n{=}25$, ASR typically stays above $0.10$, far higher than for punctuation. Human typos introduce heterogeneous and context-dependent distortions that majority voting cannot fully absorb, making them the most persistent source of errors.

\section{Discussion}
\label{sec:discussion}
Our results reveal two distinct failures in multi-agent reasoning under adversarial perturbations:

\paragraph{Semantic vs. Morphological Noise.}
We attribute the disproportionate impact of human spelling errors (\texttt{WikiTypo}) compared to punctuation noise to the degradation of semantic integrity during tokenization. Punctuation noise, while visually distracting, often preserves the morphological root of keywords, allowing the model's attention mechanism to recover the original intent. In contrast, real-world typos fundamentally alter the input's representation in the embedding space. This causes a drastic divergence from the target semantic vector, a gap that the smaller models in our study (Mistral-7B, Llama-3.1-8B) cannot bridge as the perturbed tokens might map to unrelated or out-of-vocabulary ones.

\paragraph{The Consensus Paradox.}
While classic ensemble theory assumes that agent errors are statistically independent, adversarial noise introduces a systematic bias. This bias forces the underlying model into specific, repeatable failure modes. Consequently, the agents do not exhibit the random variance required for self-correction through majority voting. Instead, they converge on the same erroneous output, causing the consensus mechanism to replicate the systematic failure rather than mitigate it. This could explain why the robustness gap remains nearly constant despite increasing $n$.

\paragraph{Trend Analysis.}
By aggregating these results, we move beyond simple repetition of dataset performance, observing that collaboration boosts the baseline accuracy of solvable problems but fails to protect the "logic gate" of the system against adversarial distortions. The consistency of this gap across six different model families suggests a fundamental limitation in how LLM agents process non-standard orthography.

\section{Conclusion}
    This work provides a systematic study of multi-LLM-agent robustness for question answering under synthetic and human-like input noise. Using a unified sampling-and-voting framework across four datasets (GSM8K, MATH, MMLU–Math, MultiArith) and six open-source models, we examined accuracy and ASR as a function of the number of collaborating agents. Three findings emerge.
    First, collaboration reliably improves accuracy, with the largest gains between one and five agents and diminishing returns beyond ten. However, ASR remains almost stable as the agent count increases. This pattern holds across models and datasets, indicating that multi-agent collaboration via majority voting is effective in improving accuracy, but the adversarial robustness gap persists.
    
    Second, the noise type matters. Punctuation corruption degrades performance roughly in proportion to its intensity, yet is largely neutralized by five to ten agents. Human typos remain the dominant bottleneck: \texttt{WikiTypo} consistently induces the largest gaps to \texttt{Clean} accuracy and the highest ASR, even at high agent counts. \texttt{R\textsuperscript{2}ATA} occupy a middle ground.
    Finally, model-level comparisons show that robustness does not strictly correlate with parameter count. While the midsize models show the highest attack success rates, both the larger and the smaller ones show more robustness. Within families such as Qwen and Gemma, larger models consistently achieve lower ASR values and thus demonstrate stronger inherent robustness. Their smaller counterparts show similar robustness while benefiting more from collaboration. Promising directions include noise-aware sampling and aggregation objectives, verifier- or tool-assisted agents, and training-time augmentation targeted at human typos. Extending beyond our scope, open-source models, and character-level corruptions to other domains, languages, and real-world pipelines will test the breadth of these gains.

\section{Limitations}
This study has several limitations. First, we restrict evaluation to English, math-reasoning benchmarks, and character-level corruptions; we do not test multilingual data, formatting/layout perturbations (e.g., LaTeX/OCR noise), or adaptive adversaries that react to model outputs. Second, our multi-agent scheme is a single instantiation-independent sampling from the same base model with fixed prompts/decoding and simple majority voting, so we do not compare against debate, tool-augmented agents, verifier-guided aggregation, diversity-promoting sampling, or cost/latency-aware strategies; the efficiency–robustness frontier therefore remains underexplored. 
Third, answer extraction is based on rule-based numeric parsing. Residual parsing errors can inflate ASR or deflate accuracy. Fourth, all agents share the same base model, so errors can be correlated, and our analysis does not disentangle correlation structure from the benefits of voting. Fifth, limited computational and financial resources constrained us from scaling the full evaluation to $n{=}100$ or sweeping various hyperparameters. However, we submit that our protocol remains robust for the claims made. A pilot study on a data subset scaling to $n{=}50$ confirmed that the adversarial robustness gap remains constant once the ensemble variance stabilizes at lower agent counts.

\section{Ethical Considerations}
\label{sec:ethics}
The persistence of the adversarial robustness gap has significant implications for the deployment of LLM agents in safety-critical, high-stakes environments. As these systems are integrated into financial modeling, medical coding, or legal reasoning, their susceptibility to simple typographical errors represents a severe reliability risk. Our findings emphasize that "scaling agents" is not a substitute for "improving base-model robustness." We highlight the necessity for character-robust tokenization or pre-processing guardrails before deploying multi-agent systems in environments where human-entered data, which is naturally noisy, is the primary input.

In addition to these safety concerns, we have conducted this research following standard ethical guidelines. We evaluate only open-source LLMs on public math benchmarks (GSM8K, MATH, MMLU–Math, MultiArith), which contain no personal or sensitive data. No human subjects were involved. Practitioners must assess risks carefully when transferring these methods beyond research. To polish the language and code, we used AI assistants, manually controlling the generated content to avoid producing harmful or biased material.

\section*{Acknowledgments}
This work was supported by the BMFTR and the state of North Rhine-Westphalia as part of the Lamarr Institute for Machine Learning and Artificial Intelligence and the AALearning Project, funded under the BMFTR grant proposal No. 05D25PD2.

\bibliography{literature}

@article{arora2025exploring,
  title={Exploring Robustness of LLMs to Paraphrasing Based on Sociodemographic Factors},
  author={Arora, Pulkit and Karimi, Akbar and Flek, Lucie},
  journal={arXiv preprint arXiv:2501.08276},
  year={2025}
}

@misc{zhao2025surveylargelanguagemodels,
      title={A Survey of Large Language Models}, 
      author={Wayne Xin Zhao and Kun Zhou and Junyi Li and Tianyi Tang and Xiaolei Wang and Yupeng Hou and Yingqian Min and Beichen Zhang and Junjie Zhang and Zican Dong and Yifan Du and Chen Yang and Yushuo Chen and Zhipeng Chen and Jinhao Jiang and Ruiyang Ren and Yifan Li and Xinyu Tang and Zikang Liu and Peiyu Liu and Jian-Yun Nie and Ji-Rong Wen},
      year={2025},
      eprint={2303.18223},
      archivePrefix={arXiv},
      primaryClass={cs.CL},
      url={https://arxiv.org/abs/2303.18223}, 
}

@misc{li2024agentsneed,
      title={More Agents Is All You Need}, 
      author={Junyou Li and Qin Zhang and Yangbin Yu and Qiang Fu and Deheng Ye},
      year={2024},
      eprint={2402.05120},
      archivePrefix={arXiv},
      primaryClass={cs.CL},
      url={https://arxiv.org/abs/2402.05120}, 
}

@article{abedin2025arithmattack,
  title={ArithmAttack: Evaluating Robustness of LLMs to Noisy Context in Math Problem Solving},
  author={Abedin, Zain Ul and Qamar, Shahzeb and Flek, Lucie and Karimi, Akbar},
  journal={arXiv preprint arXiv:2501.08203},
  year={2025},
  note={\url{https://arxiv.org/abs/2501.08203}}
}

@misc{shi2023largelanguagemodelseasily,
      title={Large Language Models Can Be Easily Distracted by Irrelevant Context}, 
      author={Freda Shi and Xinyun Chen and Kanishka Misra and Nathan Scales and David Dohan and Ed Chi and Nathanael Schärli and Denny Zhou},
      year={2023},
      eprint={2302.00093},
      archivePrefix={arXiv},
      primaryClass={cs.CL},
      url={https://arxiv.org/abs/2302.00093}, 
}

@misc{gemmateam2024gemma2improvingopen,
      title={Gemma 2: Improving Open Language Models at a Practical Size}, 
      author={{{Gemma Team}} and Morgane Riviere and Shreya Pathak and Pier Giuseppe Sessa and Cassidy Hardin and Surya Bhupatiraju and Léonard Hussenot},
      year={2024},
      eprint={2408.00118},
      archivePrefix={arXiv},
      primaryClass={cs.CL},
      url={https://arxiv.org/abs/2408.00118}, 
}

@misc{aliakbarzadeh2025exploringrobustnessmultilingualllms,
      title={Exploring Robustness of Multilingual LLMs on Real-World Noisy Data}, 
      author={Amirhossein Aliakbarzadeh and Lucie Flek and Akbar Karimi},
      year={2025},
      eprint={2501.08322},
      archivePrefix={arXiv},
      primaryClass={cs.CL},
      url={https://arxiv.org/abs/2501.08322}, 
}

@misc{gan2024reasoningrobustnessllmsadversarial,
      title={Reasoning Robustness of LLMs to Adversarial Typographical Errors}, 
      author={Esther Gan and Yiran Zhao and Liying Cheng and Yancan Mao and Anirudh Goyal and Kenji Kawaguchi and Min-Yen Kan and Michael Shieh},
      year={2024},
      eprint={2411.05345},
      archivePrefix={arXiv},
      primaryClass={cs.CL},
      url={https://arxiv.org/abs/2411.05345}, 
}

@misc{cobbe2021trainingverifierssolvemath,
      title={Training Verifiers to Solve Math Word Problems}, 
      author={Karl Cobbe and Vineet Kosaraju and Mohammad Bavarian and Mark Chen and Heewoo Jun and Lukasz Kaiser and Matthias Plappert and Jerry Tworek and Jacob Hilton and Reiichiro Nakano and Christopher Hesse and John Schulman},
      year={2021},
      eprint={2110.14168},
      archivePrefix={arXiv},
      primaryClass={cs.LG},
      url={https://arxiv.org/abs/2110.14168}, 
}

@misc{dong2023revisitinputperturbationproblems,
      title={Revisit Input Perturbation Problems for LLMs: A Unified Robustness Evaluation Framework for Noisy Slot Filling Task}, 
      author={Guanting Dong and Jinxu Zhao and Tingfeng Hui and Daichi Guo and Wenlong Wan and Boqi Feng and Yueyan Qiu and Zhuoma Gongque and Keqing He and Zechen Wang and Weiran Xu},
      year={2023},
      eprint={2310.06504},
      archivePrefix={arXiv},
      primaryClass={cs.CL},
      url={https://arxiv.org/abs/2310.06504}, 
}

@misc{yang2025qwen3technicalreport,
      title={Qwen3 Technical Report}, 
      author={An Yang and Anfeng Li and Baosong Yang and Beichen Zhang and Binyuan Hui and Bo Zheng and Bowen Yu and Chang Gao and Chengen Huang and Chenxu Lv and Chujie Zheng and Dayiheng Liu and Fan Zhou and Fei Huang and Feng Hu and Hao Ge and Haoran Wei and Huan Lin and Jialong Tang and Jian Yang and Jianhong Tu and Jianwei Zhang and Jianxin Yang and Jiaxi Yang and Jing Zhou and Jingren Zhou and Junyang Lin and Kai Dang and Keqin Bao and Kexin Yang and Le Yu and Lianghao Deng and Mei Li and Mingfeng Xue and Mingze Li and Pei Zhang and Peng Wang and Qin Zhu and Rui Men and Ruize Gao and Shixuan Liu and Shuang Luo and Tianhao Li and Tianyi Tang and Wenbiao Yin and Xingzhang Ren and Xinyu Wang and Xinyu Zhang and Xuancheng Ren and Yang Fan and Yang Su and Yichang Zhang and Yinger Zhang and Yu Wan and Yuqiong Liu and Zekun Wang and Zeyu Cui and Zhenru Zhang and Zhipeng Zhou and Zihan Qiu},
      year={2025},
      eprint={2505.09388},
      archivePrefix={arXiv},
      primaryClass={cs.CL},
      url={https://arxiv.org/abs/2505.09388}, 
}

@misc{jiang2023mistral7b,
      title={Mistral 7B}, 
      author={Albert Q. Jiang and Alexandre Sablayrolles and Arthur Mensch and Chris Bamford and Devendra Singh Chaplot and Diego de las Casas and Florian Bressand and Gianna Lengyel and Guillaume Lample and Lucile Saulnier and Lélio Renard Lavaud and Marie-Anne Lachaux and Pierre Stock and Teven Le Scao and Thibaut Lavril and Thomas Wang and Timothée Lacroix and William El Sayed},
      year={2023},
      eprint={2310.06825},
      archivePrefix={arXiv},
      primaryClass={cs.CL},
      url={https://arxiv.org/abs/2310.06825}, 
}

@misc{gemmateam2025gemma3technicalreport,
      title={Gemma 3 Technical Report}, 
      author={{{Gemma Team}} and Aishwarya Kamath and Johan Ferret and Shreya Pathak and Nino Vieillard and Ramona Merhej and Sarah Perrin and Tatiana Matejovicova and Alexandre Ramé and Morgane Rivière and Louis Rouillard and Thomas Mesnard and Geoffrey Cideron and Jean-bastien Grill and Sabela Ramos and Edouard Yvinec and Michelle Casbon and Etienne Pot and Ivo Penchev and Gaël Liu and Francesco Visin},
      year={2025},
      eprint={2503.19786},
      archivePrefix={arXiv},
      primaryClass={cs.CL},
      url={https://arxiv.org/abs/2503.19786}, 
}

@article{eger2020hero,
  title={From Hero to Z$\backslash$'eroe: A Benchmark of Low-Level Adversarial Attacks},
  author={Eger, Steffen and Benz, Yannik},
  journal={arXiv preprint arXiv:2010.05648},
  year={2020}
}

@inproceedings{formento2023using,
  title={Using punctuation as an adversarial attack on deep learning-based NLP systems: An empirical study},
  author={Formento, Brian and Foo, Chuan-Sheng and Tuan, Luu Anh and Ng, See Kiong},
  booktitle={Findings of the association for computational linguistics: EACL 2023},
  pages={1--34},
  year={2023}
}

@inproceedings{formento2021special,
  title={Special symbol attacks on nlp systems},
  author={Formento, Brian and Ng, See-Kiong and Foo, Chuan-Sheng},
  booktitle={2021 International Joint Conference on Neural Networks (IJCNN)},
  pages={1--8},
  year={2021},
  organization={IEEE}
}

@misc{lu2024blendingneedcheaperbetter,
      title={Blending Is All You Need: Cheaper, Better Alternative to Trillion-Parameters LLM}, 
      author={Xiaoding Lu and Zongyi Liu and Adian Liusie and Vyas Raina and Vineet Mudupalli and Yuwen Zhang and William Beauchamp},
      year={2024},
      eprint={2401.02994},
      archivePrefix={arXiv},
      primaryClass={cs.CL},
      url={https://arxiv.org/abs/2401.02994}, 
}

@misc{wolf2020huggingfacestransformersstateoftheartnatural,
      title={HuggingFace's Transformers: State-of-the-art Natural Language Processing}, 
      author={Thomas Wolf and Lysandre Debut and Victor Sanh and Julien Chaumond and Clement Delangue and Anthony Moi and Pierric Cistac and Tim Rault and Rémi Louf and Morgan Funtowicz and Joe Davison and Sam Shleifer and Patrick von Platen and Clara Ma and Yacine Jernite and Julien Plu and Canwen Xu and Teven Le Scao and Sylvain Gugger and Mariama Drame and Quentin Lhoest and Alexander M. Rush},
      year={2020},
      eprint={1910.03771},
      archivePrefix={arXiv},
      primaryClass={cs.CL},
      url={https://arxiv.org/abs/1910.03771}, 
}

@misc{hendrycks2021measuringmathematicalproblemsolving,
      title={Measuring Mathematical Problem Solving With the MATH Dataset}, 
      author={Dan Hendrycks and Collin Burns and Saurav Kadavath and Akul Arora and Steven Basart and Eric Tang and Dawn Song and Jacob Steinhardt},
      year={2021},
      eprint={2103.03874},
      archivePrefix={arXiv},
      primaryClass={cs.LG},
      url={https://arxiv.org/abs/2103.03874}, 
}

@misc{hendrycks2021measuringmassivemultitasklanguage,
      title={Measuring Massive Multitask Language Understanding}, 
      author={Dan Hendrycks and Collin Burns and Steven Basart and Andy Zou and Mantas Mazeika and Dawn Song and Jacob Steinhardt},
      year={2021},
      eprint={2009.03300},
      archivePrefix={arXiv},
      primaryClass={cs.CY},
      url={https://arxiv.org/abs/2009.03300}, 
}

@misc{roy2016solvinggeneralarithmeticword,
      title={Solving General Arithmetic Word Problems}, 
      author={Subhro Roy and Dan Roth},
      year={2016},
      eprint={1608.01413},
      archivePrefix={arXiv},
      primaryClass={cs.CL},
      url={https://arxiv.org/abs/1608.01413}, 
}

@article{llamatechnicalreport,
  title={The llama 3 herd of models},
  author={Dubey, Abhimanyu and Jauhri, Abhinav and Pandey, Abhinav and Kadian, Abhishek and Al-Dahle, Ahmad and Letman, Aiesha and Mathur, Akhil and Schelten, Alan and Yang, Amy and Fan, Angela and others},
  journal={arXiv e-prints},
  pages={arXiv--2407},
  year={2024}
}

@inproceedings{kwon2023efficient,
  title={Efficient Memory Management for Large Language Model Serving with PagedAttention},
  author={Woosuk Kwon and Zhuohan Li and Siyuan Zhuang and Ying Sheng and Lianmin Zheng and Cody Hao Yu and Joseph E. Gonzalez and Hao Zhang and Ion Stoica},
  booktitle={Proceedings of the ACM SIGOPS 29th Symposium on Operating Systems Principles},
  year={2023}
}

@misc{helmi2025modelingresponseconsistencymultiagent,
      title={Modeling Response Consistency in Multi-Agent LLM Systems: A Comparative Analysis of Shared and Separate Context Approaches}, 
      author={Tooraj Helmi},
      year={2025},
      eprint={2504.07303},
      archivePrefix={arXiv},
      primaryClass={cs.MA},
      url={https://arxiv.org/abs/2504.07303}, 
}

@INPROCEEDINGS{gao2018black,
  author={Gao, Ji and Lanchantin, Jack and Soffa, Mary Lou and Qi, Yanjun},
  booktitle={2018 IEEE Security and Privacy Workshops (SPW)}, 
  title={Black-Box Generation of Adversarial Text Sequences to Evade Deep Learning Classifiers}, 
  year={2018},
  volume={},
  number={},
  pages={50-56},
  doi={10.1109/SPW.2018.00016}}

@article{li2018textbugger,
  title={Textbugger: Generating adversarial text against real-world applications},
  author={Li, Jinfeng and Ji, Shouling and Du, Tianyu and Li, Bo and Wang, Ting},
  journal={arXiv preprint arXiv:1812.05271},
  year={2018}
}

@inproceedings{garg2020bae,
    title = "{BAE}: {BERT}-based Adversarial Examples for Text Classification",
    author = "Garg, Siddhant  and
      Ramakrishnan, Goutham",
    editor = "Webber, Bonnie  and
      Cohn, Trevor  and
      He, Yulan  and
      Liu, Yang",
    booktitle = "Proceedings of the 2020 Conference on Empirical Methods in Natural Language Processing (EMNLP)",
    month = nov,
    year = "2020",
    address = "Online",
    publisher = "Association for Computational Linguistics",
    url = "https://aclanthology.org/2020.emnlp-main.498/",
    doi = "10.18653/v1/2020.emnlp-main.498",
    pages = "6174--6181"
}

@inproceedings{moradi2021evaluating,
    title = "Evaluating the Robustness of Neural Language Models to Input Perturbations",
    author = "Moradi, Milad  and
      Samwald, Matthias",
    editor = "Moens, Marie-Francine  and
      Huang, Xuanjing  and
      Specia, Lucia  and
      Yih, Scott Wen-tau",
    booktitle = "Proceedings of the 2021 Conference on Empirical Methods in Natural Language Processing",
    month = nov,
    year = "2021",
    address = "Online and Punta Cana, Dominican Republic",
    publisher = "Association for Computational Linguistics",
    url = "https://aclanthology.org/2021.emnlp-main.117/",
    doi = "10.18653/v1/2021.emnlp-main.117",
    pages = "1558--1570"
}

@InProceedings{shi2023robustness,
  title = 	 {Large Language Models Can Be Easily Distracted by Irrelevant Context},
  author =       {Shi, Freda and Chen, Xinyun and Misra, Kanishka and Scales, Nathan and Dohan, David and Chi, Ed H. and Sch\"{a}rli, Nathanael and Zhou, Denny},
  booktitle = 	 {Proceedings of the 40th International Conference on Machine Learning},
  pages = 	 {31210--31227},
  year = 	 {2023},
  editor = 	 {Krause, Andreas and Brunskill, Emma and Cho, Kyunghyun and Engelhardt, Barbara and Sabato, Sivan and Scarlett, Jonathan},
  volume = 	 {202},
  series = 	 {Proceedings of Machine Learning Research},
  month = 	 {23--29 Jul},
  publisher =    {PMLR},
  url = 	 {https://proceedings.mlr.press/v202/shi23a.html}
}

@misc{lanham2023adversarial,
      title={Measuring Faithfulness in Chain-of-Thought Reasoning}, 
      author={Tamera Lanham and Anna Chen and Ansh Radhakrishnan and Benoit Steiner and Carson Denison and Danny Hernandez and Dustin Li and Esin Durmus and Evan Hubinger and Jackson Kernion and Kamilė Lukošiūtė and Karina Nguyen and Newton Cheng and Nicholas Joseph and Nicholas Schiefer and Oliver Rausch and Robin Larson and Sam McCandlish and Sandipan Kundu and Saurav Kadavath and Shannon Yang and Thomas Henighan and Timothy Maxwell and Timothy Telleen-Lawton and Tristan Hume and Zac Hatfield-Dodds and Jared Kaplan and Jan Brauner and Samuel R. Bowman and Ethan Perez},
      year={2023},
      eprint={2307.13702},
      archivePrefix={arXiv},
      primaryClass={cs.AI},
      url={https://arxiv.org/abs/2307.13702}, 
}

@misc{wang2025tdagmultiagentframeworkbased,
      title={TDAG: A Multi-Agent Framework based on Dynamic Task Decomposition and Agent Generation}, 
      author={Yaoxiang Wang and Zhiyong Wu and Junfeng Yao and Jinsong Su},
      year={2025},
      eprint={2402.10178},
      archivePrefix={arXiv},
      primaryClass={cs.CL},
      url={https://arxiv.org/abs/2402.10178}, 
}

@misc{talebirad2023multiagentcollaborationharnessingpower,
      title={Multi-Agent Collaboration: Harnessing the Power of Intelligent LLM Agents}, 
      author={Yashar Talebirad and Amirhossein Nadiri},
      year={2023},
      eprint={2306.03314},
      archivePrefix={arXiv},
      primaryClass={cs.AI},
      url={https://arxiv.org/abs/2306.03314}, 
}

@misc{seo2025spioensembleselectivestrategies,
      title={SPIO: Ensemble and Selective Strategies via LLM-Based Multi-Agent Planning in Automated Data Science}, 
      author={Wonduk Seo and Juhyeon Lee and Yi Bu},
      year={2025},
      eprint={2503.23314},
      archivePrefix={arXiv},
      primaryClass={cs.AI},
      url={https://arxiv.org/abs/2503.23314}, 
}

@inproceedings{becker2025mallm,
  title={Mallm: Multi-agent large language models framework},
  author={Becker, Jonas and others},
  booktitle={Proceedings of the 2025 Conference on Empirical Methods in Natural Language Processing: System Demonstrations},
  year={2025}
}

@article{asad2025reddebate,
  title={RedDebate: Safer Responses through Multi-Agent Red Teaming Debates},
  author={Asad, Ali and others},
  journal={arXiv preprint arXiv:2506.11083},
  year={2025}
}

@inproceedings{li2024improving,
  title={Improving Multi-Agent Debate with Sparse Communication Topology},
  author={Li, Yunxuan and Du, Yibing and Zhang, Jiageng and Hou, Le
          and Grabowski, Peter and Li, Yeqing and Ie, Eugene},
  booktitle={Findings of the Association for Computational Linguistics: EMNLP 2024},
  pages={7281--7294},
  year={2024}
}

@article{tran2025survey,
  title={Multi-Agent Collaboration Mechanisms: A Survey of {LLMs}},
  author={Tran, Khanh-Tung and Dao, Dung and Nguyen, Minh-Duong
          and Pham, Quoc-Viet and O'Sullivan, Barry and Nguyen, Hoang D.},
  journal={arXiv preprint arXiv:2501.06322},
  year={2025}
}

@inproceedings{du2023improving,
  title={Improving Factuality and Reasoning in Language Models
         through Multiagent Debate},
  author={Du, Yilun and Li, Shuang and Torralba, Antonio
          and Tenenbaum, Joshua B. and Mordatch, Igor},
  booktitle={Proceedings of the 41st International Conference
             on Machine Learning},
  year={2024}
}

\appendix

\section{Qualitative Failure Cases}

The following are some of the failure cases under different attacks: 
\\

\noindent 
\begin{tcolorbox}[
  colback=green!5!white,
  colframe=green!40!black,
  colbacktitle=green!40!black,
  title=Dataset: MATH,
  breakable,        
  enhanced,         
  boxrule=0.6pt,
  arc=2pt,
  before upper={
    \textbf{Noise type: punctuation\_10} \\
    \textbf{Ground truth: $\frac{9}{4},-\frac{9}{4}$} \\
    \\
  \textbf{Question:}\par\smallskip},
  after title={\par\smallskip}
]
Here is a math problem written in LaTeX: Find all real numbers $k$ such that $x^4+kx^3+x^2+4kx+16=0$ is true for exactly , one real number $x = r$. Enter all the possible values of ! $k,$ separated by commas.

Solve it step by step. 
- If the correct answer has units, write the units OUTSIDE the box in plain text. 
- Inside \verb|\boxed{...}| put only the Latex style exact numeric expression (with units) 
(e.g., rationalized denominator, reduced fractions, simplified radicals). 
At the very end of your response, output exactly one line: \verb|\boxed{<simplest_exact_value>}| 
If units are required, add as a \verb|\text{ unit }| and then the units in words after the box, e.g.: \verb|\boxed{ 12 \text{ square inches }}| 
Do not add any extra commentary after that line.

\vspace{0.8em}
\tcblower
\vspace{0.8em}
\noindent\textbf{Output:} \\

Let $P(x) = x^4+kx^3+x^2+4kx+16$. 
We require $P(x) = 0$ to have exactly one real root $x=r$. 
We can rewrite $P(x)$ as:
$P(x) = x^4+x^2+16 + kx^3+4kx = (x^2+1)^2 + kx(x^2+4) = 0$.

If $k=0$, then $P(x) = x^4+x^2+16$. 
The discriminant of $x^4+x^2+16$ is $16-4(16)=16-64 = -48 < 0$. 
Since the leading coefficient is positive, $P(x) > 0$ for all real $x$. Therefore, $P(x)=0$ has no real roots.

Consider $k \ne 0$. \textcolor{red}{Suppose $r$ is a repeated root of $P(x)$. Then $P(r) = 0$ and $P'(r) = 0$.} 
$P'(x) = 4x^3+3kx^2+2x+4k$. 
$P(r) = r^4+kr^3+r^2+4kr+16 = 0$
$P'(r) = 4r^3+3kr^2+2r+4k = 0$
$4r^3+3kr^2+2r+4k = 0 \Rightarrow k(3r^2+4) = -4r^3-2r$
If $3r^2+4 \ne 0$, then $k = \frac{-4r^3-2r}{3r^2+4}$.
Substituting this into $P(r)=0$, we get 
$r^4 + \frac{-4r^3-2r}{3r^2+4} r^3 + r^2+4 \frac{-4r^3-2r}{3r^2+4} r + 16 = 0$
$r^4 + \frac{-4r^6-2r^4}{3r^2+4} + r^2 + \frac{-16r^4-8r^2}{3r^2+4} + 16 = 0$
Multiplying by $3r^2+4$, we get
$r^4(3r^2+4) - 4r^6-2r^4 + r^2(3r^2+4) - 16r^4-8r^2 + 16(3r^2+4) = 0$
$3r^6+4r^4-4r^6-2r^4+3r^4+4r^2-16r^4-8r^2+48r^2+64 = 0$
$-r^6 - 11r^4+44r^2+64=0$
$r^6+11r^4-44r^2-64=0$
Let $y = r^2$. Then $y^3+11y^2-44y-64=0$. 
Since $r$ is real, $y=r^2 \ge 0$. 

However, we are seeking for exactly one real root. 
Consider the case where $x^4+kx^3+x^2+4kx+16 = (x^2+ax+b)(x^2+cx+d)$ where $a, b, c, d$ are real numbers. 
Since $P(x)$ has degree 4, we have $x^4+kx^3+x^2+4kx+16 = (x^2+ax+b)(x^2+cx+d)$ 
$= x^4+(a+c)x^3 + (ac+b+d)x^2+(ad+bc)x+bd$ 
Comparing coefficients, we have 
$a+c = k$ 
$ac+b+d = 1$ 
$ad+bc = 4k$ 
$bd = 16$ 
If $b=d=4$, then $ac+8 = 1$, so $ac = -7$. 
$4a+4c = 4k$, so $a+c = k$. 
So $a+c = k$ and $ac = -7$. 
$a$ and $c$ are roots of the equation $t^2-kt-7=0$. 
The discriminant is $k^2+28$. 
For $a, c$ to be real, $k^2+28 \ge 0$, which is always true. 
$t = \frac{k \pm \sqrt{k^2+28}}{2}$. 
Thus $a = \frac{k + \sqrt{k^2+28}}{2}$ and $c = \frac{k - \sqrt{k^2+28}}{2}$. 
Then $P(x) = (x^2+(\frac{k + \sqrt{k^2+28}}{2})x + 4)(x^2+(\frac{k - \sqrt{k^2+28}}{2})x + 4)$ 
We want exactly one real root. We need $x^2+(\frac{k - \sqrt{k^2+28}}{2})x + 4 = 0$ to have no real roots, so 
$(\frac{k - \sqrt{k^2+28}}{2})^2 - 4 \cdot 1 < 0$. 
$\frac{k^2 - 2k\sqrt{k^2+28} + k^2+28}{4} - 4 < 0$ 
$2k^2+28 - 2k\sqrt{k^2+28} < -16$ 
$2k^2+44 < 2k\sqrt{k^2+28}$ 
$k^2+22 < k\sqrt{k^2+28}$ 
Squaring both sides: $(k^2+22)^2 < k^2(k^2+28)$ 
$k^4+44k^2+484 < k^4+28k^2$ 
$16k^2 < -484$, which is impossible since $k$ is real. 

If $b=d=16$, then $ac+16+16 = 1$, so $ac = -31$. 
$16a+16c = 4k$, so $a+c = \frac{1}{4}k$. 
Then $a$ and $c$ are roots of $t^2-\frac{k}{4}t-31=0$. 
$t = \frac{\frac{k}{4} \pm \sqrt{\frac{k^2}{16}+124}}{2} = \frac{k \pm \sqrt{k^2+1984}}{8}$. 
Then we require $\frac{k^2}{16} + 124 < 0$, which is impossible. 

Let $k=0$. Then $x^4+x^2+16=0$. This has no real roots. 
Let $P(x)=0$. If $P(x)$ has only one real root then $P'(x)=0$ at that root. Thus $x=r$ must be a double root. 
Then $P(r)=0$ and $P'(r)=0$. 

Let $k=2\sqrt{2}$. $P(x) = x^4+2\sqrt{2}x^3+x^2+4\sqrt{2}x+16$. 

Final Answer: The final answer is $\boxed{0}$
\end{tcolorbox}
\noindent\textbf{Takeaway:} This example shows how punctuation noise can disrupt the model's reasoning in complex mathematical problems. The model incorrectly assumes that exactly one real root must be a repeated root, leading to a flawed approach and incorrect conclusion. This highlights that noise can cause the model to make invalid assumptions and fail to apply correct mathematical principles.
\\

\noindent 
\begin{tcolorbox}[
  colback=green!5!white,
  colframe=green!40!black,
  colbacktitle=green!40!black,
  title=Dataset: GSM8K,
  breakable,        
  enhanced,         
  boxrule=0.6pt,
  arc=2pt,
  before upper={
    \textbf{Noise type: Wikitypo} \\
    \textbf{Ground truth: $18$} \\
    \\
  \textbf{Question:}\par\smallskip},
  after title={\par\smallskip}
]

Can you solve the following math problem? Janet’s ducks lay 16 eggs por day. The eats three \textcolor{red}{por} breakfast every morning Rand bakes muffins \textcolor{red}{por} heir friends every days \textcolor{red}{Width} four. The sells Then remainder \textcolor{red}{nat} Then farmers' markets daily \textcolor{red}{por} \$2 \textcolor{red}{por} fresh dock egg. How much \textcolor{red}{im} dollars does The made every days \textcolor{red}{nat Then} farmers' market? Explain your reasoning. Your final answer should be a single numerical number, in the form \boxed{answer}, at the end of your response. 

  \tcblower
  \noindent\textbf{Output:} \\ 
Here's how to break down the problem and solve it step-by-step:
1. **Eggs laid per day:** Janet's ducks lay 16 eggs per day.
2. \textcolor{red}{**Eggs eaten:** Each day, the ducks eat 3 eggs.}
3. **Eggs remaining after being eaten:** 16 eggs - 3 eggs = 13 eggs
4. **Muffins baked per day:** Janet bakes 4 muffins per day. This information is irrelevant to the final result.
5. **Eggs sold:**  Janet sells the remaining 13 eggs.
6. **Income from egg sales:** She sells each egg for \$2, so she makes 13 eggs * \$2/egg = \$26.
7. **Total earnings:** Janet makes \$26 per day at the farmers' market.
\boxed{26}
\end{tcolorbox}
\noindent\textbf{Takeaway:}
This example illustrates how Wikitypo noise (intentional typos/misspellings) can cause the model to overlook critical arithmetic operations. The noisy text changed "eats three for breakfast" to "eats three por breakfast" and introduced other distracting misspellings like "por" instead of "for," "nat" instead of "at," and "im" instead of "in." While the model correctly extracted the key numbers (16 eggs, 3 eaten, \$2 per egg), it failed to recognize that the muffins information was entirely irrelevant noise and incorrectly performed the subtraction (16-3=13) to get \$26, when the correct interpretation should have been that all 16 eggs are sold directly without any subtraction for breakfast consumption. The typos obscured the logical relationships between quantities, leading to an incorrect arithmetic operation and final answer.
\\


\noindent 
\begin{tcolorbox}[
  colback=green!5!white,
  colframe=green!40!black,
  colbacktitle=green!40!black,
  title=Dataset: MultiArith,
  breakable,        
  enhanced,         
  boxrule=0.6pt,
  arc=2pt,
  before upper={
    \textbf{Noise type: punctuation\_50} \\
    \textbf{Ground truth: $72$} \\
    \\
  \textbf{Question:}\par\smallskip},
  after title={\par\smallskip}
]
Can you solve the following math problem? Megan . ; earned ? ; ? 8 . points ? for each bag of cans she recycled. . If . , she had , 14 bags, but didn't recycle 5 : of them, how ; many points ! would she have earned? Explain your reasoning. Your final answer should be a single numerical number, in the form \boxed{answer}, at the end of your response. 

  \tcblower
  \noindent\textbf{Output:} \\ 
   First, let's figure out the points Megan earned for recycling each bag of cans. She earned 8 points for each bag. \\
   Maybe we can add that up for all the bags she recycled. That would be 8 points / bag times 14 bags = \boxed{112} \\
   However, she didn't recycle 5 bags, \textcolor{red}{so we can't add anything for those un-recycled bags.} \\
   Therefore, she would have earned \boxed{112} points before she recycled only 14 bags of cans.

\end{tcolorbox}

\noindent\textbf{Takeaway:} This example demonstrates how inserting punctuation noise can lead to task misunderstanding. The original problem intended that Megan earned 112 points for 14 bags, it noted the 5 un-recycled bags but the model failed to subtract them from the total, resulting in an incorrect calculation.
\\

\noindent 
\begin{tcolorbox}[
  colback=green!5!white,
  colframe=green!40!black,
  colbacktitle=green!40!black,
  title=Dataset: MMLU,
  breakable,        
  enhanced,         
  boxrule=0.6pt,
  arc=2pt,
  before upper={
    \textbf{Noise type: R\textsuperscript{2}{ATA}} \\
    \textbf{Ground truth: $(B)$} \\
    \\
  \textbf{Question:}\par\smallskip},
  after title={\par\smallskip}
]
Can you answer the following question as accurately as possible? Question: Statement 1 | Any set of ttwo vectors in $R^2$ is linearly indrpendent. Statement 2 | If $V = span(v1,..., vk)$ ajd ${v1,..., vk}$ are linearly independent, then $dim(V) = k.$ : 
 Choices: A) True, True, B) False, False, C) True, False, D) False, True. Briefly explain your reasoning. Choose only one of A, B, C, or D. At the end, output only one final answer like \boxed{(A)}, \boxed{(B)}, \boxed{(C)}, or \boxed{(D)}. Do not add extra commentary after the answer. Do not include anything else on that final line.

  \tcblower
  \noindent\textbf{Output:} \\ 
Statement 1 is False. In $R^2$, we can find two vectors that are linearly dependent (e.g., $v1 = (1, 0)$ and $v2 = (2, 0))$. \textcolor{red}{Statement 2 is True. If $V = span(v1, ..., vk)$ and ${v1, ..., vk}$ are linearly independent, then the vectors $v1, ..., vk$ form a basis for $V$, and the dimension of $V$ is equal to the number of vectors in the set, which is $k$.}
\boxed{(B)}

\end{tcolorbox}
\noindent\textbf{Takeaway:} This example demonstrates how character substitution noise (\texttt{R\textsuperscript{2}{ATA}}) can lead to logical inconsistencies in the model's reasoning. The model correctly identifies Statement 1 as false, but then contradicts itself by stating Statement 2 is true while selecting option (B) which requires both statements to be false, showing confusion in maintaining logical consistency throughout the response.

\clearpage
\onecolumn
\section{Extended Results}\label{extended_results}

\begin{figure}[t]
    \centering
    
    \includegraphics[width=0.47\textwidth]{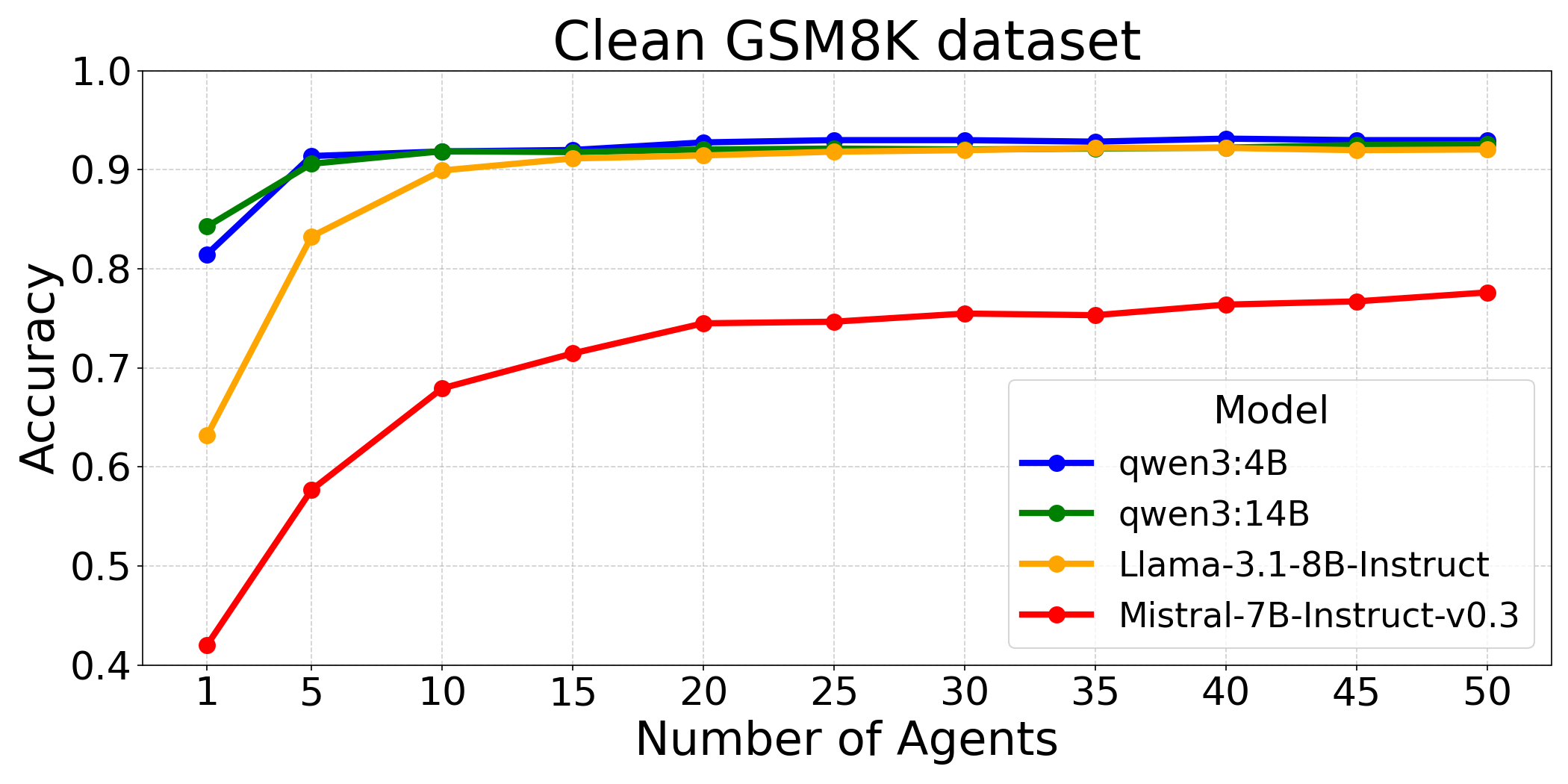}
    \caption{Model accuracy trend by the number of agents (Noise type: clean). The graph shows the performance of four different models (qwen3:4B, qwen3:14B, Llama-3.1-8B-Instruct, and Mistral-7B-Instruct-v0.3) as the number of agents increases from 1 to 50. All models demonstrate improved accuracy with more agents, with qwen3:14B achieving the highest performance, followed closely by qwen3:4B and Llama-3.1-8B-Instruct. Mistral-7B-Instruct-v0.3 shows lower overall accuracy but still benefits from the multi-agent approach.}
    \label{fig:appendix_model_acc_50}
\end{figure}

\begin{table*}[t]
\centering
\resizebox{\textwidth}{!}{%
\begin{tabular}{l c c c c c c c|c}
\hline
Model & Agent Count & Clean & Punct-10 & Punct-30 & Punct-50 & WikiTypo & R\textsuperscript{2}ATA & Avg \\
\hline
Llama-3.1 
 & 1  & 0.532 $\pm$ 0.192 & 0.532 $\pm$ 0.213 & 0.515 $\pm$ 0.204 & 0.505 $\pm$ 0.206 & 0.458 $\pm$ 0.197 & 0.540 $\pm$ 0.208 & 0.514 \\
 & 5  & 0.688 $\pm$ 0.219 & 0.666 $\pm$ 0.234 & 0.641 $\pm$ 0.227 & 0.626 $\pm$ 0.220 & 0.586 $\pm$ 0.232 & 0.671 $\pm$ 0.227 & 0.646 \\
 & 10 & 0.752 $\pm$ 0.205 & 0.725 $\pm$ 0.220 & 0.695 $\pm$ 0.224 & 0.686 $\pm$ 0.211 & 0.644 $\pm$ 0.229 & 0.724 $\pm$ 0.212 & 0.704 \\
 & 20 & 0.775 $\pm$ 0.192 & 0.747 $\pm$ 0.207 & 0.718 $\pm$ 0.213 & 0.709 $\pm$ 0.209 & 0.673 $\pm$ 0.219 & 0.753 $\pm$ 0.200 & 0.729 \\
\hline
Mistral-7B
 & 1  & 0.395 $\pm$ 0.228 & 0.384 $\pm$ 0.233 & 0.360 $\pm$ 0.213 & 0.340 $\pm$ 0.203 & 0.339 $\pm$ 0.242 & 0.376 $\pm$ 0.228 & 0.366 \\
 & 5  & 0.497 $\pm$ 0.292 & 0.480 $\pm$ 0.297 & 0.457 $\pm$ 0.274 & 0.431 $\pm$ 0.271 & 0.425 $\pm$ 0.307 & 0.465 $\pm$ 0.287 & 0.459 \\
 & 10 & 0.553 $\pm$ 0.298 & 0.526 $\pm$ 0.311 & 0.512 $\pm$ 0.287 & 0.489 $\pm$ 0.287 & 0.470 $\pm$ 0.320 & 0.517 $\pm$ 0.297 & 0.511 \\
 & 20 & 0.578 $\pm$ 0.301 & 0.556 $\pm$ 0.316 & 0.543 $\pm$ 0.291 & 0.517 $\pm$ 0.303 & 0.498 $\pm$ 0.312 & 0.543 $\pm$ 0.302 & 0.539 \\
\hline
Qwen3-14B
 & 1  & 0.800 $\pm$ 0.172 & 0.782 $\pm$ 0.186 & 0.779 $\pm$ 0.188 & 0.777 $\pm$ 0.184 & 0.740 $\pm$ 0.192 & 0.785 $\pm$ 0.185 & 0.777 \\
 & 5  & 0.846 $\pm$ 0.138 & 0.824 $\pm$ 0.157 & 0.826 $\pm$ 0.156 & 0.824 $\pm$ 0.160 & 0.786 $\pm$ 0.165 & 0.821 $\pm$ 0.160 & 0.821 \\
 & 10 & 0.863 $\pm$ 0.116 & 0.847 $\pm$ 0.128 & 0.844 $\pm$ 0.132 & 0.844 $\pm$ 0.138 & 0.802 $\pm$ 0.148 & 0.836 $\pm$ 0.142 & 0.839 \\
 & 20 & 0.880 $\pm$ 0.091 & 0.864 $\pm$ 0.105 & 0.855 $\pm$ 0.120 & 0.859 $\pm$ 0.118 & 0.820 $\pm$ 0.125 & 0.856 $\pm$ 0.117 & 0.855 \\
\hline
Qwen3-4B
 & 1  & 0.786 $\pm$ 0.164 & 0.757 $\pm$ 0.170 & 0.741 $\pm$ 0.157 & 0.748 $\pm$ 0.159 & 0.679 $\pm$ 0.210 & 0.749 $\pm$ 0.164 & 0.743 \\
 & 5  & 0.848 $\pm$ 0.127 & 0.831 $\pm$ 0.142 & 0.816 $\pm$ 0.135 & 0.813 $\pm$ 0.140 & 0.749 $\pm$ 0.171 & 0.820 $\pm$ 0.135 & 0.813 \\
 & 10 & 0.864 $\pm$ 0.114 & 0.852 $\pm$ 0.126 & 0.836 $\pm$ 0.116 & 0.835 $\pm$ 0.123 & 0.774 $\pm$ 0.148 & 0.838 $\pm$ 0.117 & 0.833 \\
 & 20 & 0.876 $\pm$ 0.103 & 0.865 $\pm$ 0.110 & 0.848 $\pm$ 0.105 & 0.849 $\pm$ 0.106 & 0.787 $\pm$ 0.135 & 0.855 $\pm$ 0.096 & 0.847 \\
\hline
gemma-3-12b
 & 1  & 0.841 $\pm$ 0.147 & 0.835 $\pm$ 0.139 & 0.831 $\pm$ 0.134 & 0.818 $\pm$ 0.136 & 0.771 $\pm$ 0.160 & 0.834 $\pm$ 0.147 & 0.822 \\
 & 5  & 0.862 $\pm$ 0.134 & 0.856 $\pm$ 0.130 & 0.855 $\pm$ 0.125 & 0.838 $\pm$ 0.131 & 0.796 $\pm$ 0.149 & 0.855 $\pm$ 0.132 & 0.843 \\
 & 10 & 0.868 $\pm$ 0.126 & 0.860 $\pm$ 0.129 & 0.861 $\pm$ 0.118 & 0.844 $\pm$ 0.127 & 0.803 $\pm$ 0.144 & 0.862 $\pm$ 0.125 & 0.850 \\
 & 20 & \textbf{0.874 $\pm$ 0.119} & 0.861 $\pm$ 0.129 & \textbf{0.866 $\pm$ 0.115} & 0.846 $\pm$ 0.130 & 0.809 $\pm$ 0.139 & \textbf{0.866 $\pm$ 0.121} & 0.854 \\
\hline
gemma-3-4b
 & 1  & 0.771 $\pm$ 0.161 & 0.745 $\pm$ 0.166 & 0.733 $\pm$ 0.171 & 0.725 $\pm$ 0.174 & 0.651 $\pm$ 0.207 & 0.731 $\pm$ 0.184 & 0.726 \\
 & 5  & 0.822 $\pm$ 0.138 & 0.792 $\pm$ 0.144 & 0.787 $\pm$ 0.148 & 0.777 $\pm$ 0.149 & 0.705 $\pm$ 0.184 & 0.782 $\pm$ 0.151 & 0.778 \\
 & 10 & 0.837 $\pm$ 0.125 & 0.805 $\pm$ 0.134 & 0.800 $\pm$ 0.140 & 0.786 $\pm$ 0.146 & 0.718 $\pm$ 0.176 & 0.800 $\pm$ 0.131 & 0.791 \\
 & 20 & 0.845 $\pm$ 0.118 & \textbf{0.865 $\pm$ 0.110} & 0.810 $\pm$ 0.131 & \textbf{0.859 $\pm$ 0.118} & \textbf{0.820 $\pm$ 0.125} & 0.806 $\pm$ 0.127 & 0.797 \\
\hline
\textbf{Average}
 & 1  & 0.688 $\pm$ 0.243 & 0.672 $\pm$ 0.246 & 0.660 $\pm$ 0.245 & 0.652 $\pm$ 0.248 & 0.606 $\pm$ 0.256 & 0.669 $\pm$ 0.247 & 0.658 \\
 & 5  & 0.760 $\pm$ 0.227 & 0.741 $\pm$ 0.235 & 0.730 $\pm$ 0.233 & 0.718 $\pm$ 0.236 & 0.674 $\pm$ 0.246 & 0.736 $\pm$ 0.233 & 0.727 \\
 & 10 & 0.789 $\pm$ 0.210 & 0.769 $\pm$ 0.222 & 0.758 $\pm$ 0.219 & 0.747 $\pm$ 0.222 & 0.702 $\pm$ 0.236 & 0.763 $\pm$ 0.218 & 0.755 \\
 & 20 & 0.805 $\pm$ 0.202 & 0.793 $\pm$ 0.213 & 0.773 $\pm$ 0.210 & 0.773 $\pm$ 0.219 & 0.735 $\pm$ 0.222 & 0.780 $\pm$ 0.209 & 0.770 \\
\hline
\end{tabular}
} 
\caption{Averaged accuracy (mean $\pm$ std) across datasets for each model and agent count under different noise types. Larger models (e.g., Qwen3-14B, Gemma3-12B) achieve the highest overall accuracies, while smaller variants (Qwen3-4B, Gemma3-4B) show steeper gains with increasing agent count. Mistral-7B remains lowest across all settings, highlighting model-specific limitations. Bold values indicate the best accuracy per column.}
\label{tab:accuracy_results}
\end{table*}

\begin{table*}[t]
\centering
\resizebox{\textwidth}{!}{%
\begin{tabular}{l c c c c c c c|c}
\hline
\textbf{Model} & \textbf{Agent Count} & \textbf{Clean} & \textbf{Punct-10} & \textbf{Punct-30} & \textbf{Punct-50} & \textbf{WikiTypo} & \textbf{R\textsuperscript{2}ATA} & \textbf{Avg} \\
\hline
\textbf{Llama-3.1} 
 & 1  & 0.518 $\pm$ 0.015 & 0.490 $\pm$ 0.016 & 0.469 $\pm$ 0.014 & 0.451 $\pm$ 0.013 & 0.441 $\pm$ 0.014 & 0.501 $\pm$ 0.015 & 0.478 \\
 & 5  & 0.623 $\pm$ 0.011 & 0.588 $\pm$ 0.012 & 0.574 $\pm$ 0.008 & 0.553 $\pm$ 0.010 & 0.537 $\pm$ 0.010 & 0.603 $\pm$ 0.010 & 0.580 \\
 & 10 & 0.662 $\pm$ 0.010 & 0.630 $\pm$ 0.011 & 0.620 $\pm$ 0.008 & 0.597 $\pm$ 0.007 & 0.575 $\pm$ 0.008 & 0.639 $\pm$ 0.009 & 0.621 \\
 & 15 & 0.679 $\pm$ 0.004 & 0.650 $\pm$ 0.010 & 0.626 $\pm$ 0.005 & 0.611 $\pm$ 0.009 & 0.591 $\pm$ 0.004 & 0.656 $\pm$ 0.005 & 0.636 \\
 & 20 & 0.683 $\pm$ 0.003 & 0.658 $\pm$ 0.007 & 0.635 $\pm$ 0.004 & 0.621 $\pm$ 0.004 & 0.603 $\pm$ 0.004 & 0.665 $\pm$ 0.004 & 0.644 \\
 & 25 & 0.686 $\pm$ 0.000 & 0.663 $\pm$ 0.000 & 0.635 $\pm$ 0.000 & 0.624 $\pm$ 0.000 & 0.607 $\pm$ 0.000 & 0.672 $\pm$ 0.000 & 0.648 \\
\hline
\textbf{Mistral-7B}
 & 1  & 0.374 $\pm$ 0.013 & 0.369 $\pm$ 0.015 & 0.358 $\pm$ 0.012 & 0.350 $\pm$ 0.010 & 0.339 $\pm$ 0.013 & 0.370 $\pm$ 0.015 & 0.360 \\
 & 5  & 0.421 $\pm$ 0.010 & 0.418 $\pm$ 0.012 & 0.405 $\pm$ 0.010 & 0.394 $\pm$ 0.009 & 0.380 $\pm$ 0.010 & 0.418 $\pm$ 0.009 & 0.406 \\
 & 10 & 0.442 $\pm$ 0.006 & 0.431 $\pm$ 0.006 & 0.421 $\pm$ 0.008 & 0.416 $\pm$ 0.009 & 0.395 $\pm$ 0.008 & 0.437 $\pm$ 0.006 & 0.424 \\
 & 15 & 0.448 $\pm$ 0.004 & 0.437 $\pm$ 0.005 & 0.430 $\pm$ 0.005 & 0.425 $\pm$ 0.007 & 0.402 $\pm$ 0.006 & 0.444 $\pm$ 0.006 & 0.431 \\
 & 20 & 0.449 $\pm$ 0.006 & 0.444 $\pm$ 0.005 & 0.432 $\pm$ 0.005 & 0.420 $\pm$ 0.003 & 0.403 $\pm$ 0.003 & 0.448 $\pm$ 0.006 & 0.433 \\
 & 25 & 0.455 $\pm$ 0.000 & 0.449 $\pm$ 0.000 & 0.437 $\pm$ 0.000 & 0.426 $\pm$ 0.000 & 0.411 $\pm$ 0.000 & 0.447 $\pm$ 0.000 & 0.437 \\
\hline
\textbf{Qwen3-14B}
 & 1  & 0.822 $\pm$ 0.007 & 0.814 $\pm$ 0.006 & 0.808 $\pm$ 0.006 & 0.804 $\pm$ 0.007 & 0.788 $\pm$ 0.009 & 0.807 $\pm$ 0.007 & 0.807 \\
 & 5  & 0.836 $\pm$ 0.005 & 0.828 $\pm$ 0.005 & 0.822 $\pm$ 0.006 & 0.823 $\pm$ 0.004 & 0.803 $\pm$ 0.006 & 0.823 $\pm$ 0.004 & 0.822 \\
 & 10 & 0.836 $\pm$ 0.003 & 0.830 $\pm$ 0.004 & 0.821 $\pm$ 0.004 & 0.824 $\pm$ 0.004 & 0.806 $\pm$ 0.004 & 0.824 $\pm$ 0.003 & 0.824 \\
 & 15 & 0.833 $\pm$ 0.002 & 0.829 $\pm$ 0.003 & 0.822 $\pm$ 0.002 & 0.823 $\pm$ 0.003 & 0.806 $\pm$ 0.005 & 0.825 $\pm$ 0.004 & 0.823 \\
 & 20 & 0.834 $\pm$ 0.003 & 0.829 $\pm$ 0.002 & 0.825 $\pm$ 0.002 & 0.823 $\pm$ 0.003 & 0.804 $\pm$ 0.002 & 0.826 $\pm$ 0.001 & 0.824 \\
 & 25 & 0.832 $\pm$ 0.000 & 0.830 $\pm$ 0.000 & 0.830 $\pm$ 0.000 & 0.821 $\pm$ 0.000 & 0.805 $\pm$ 0.000 & 0.827 $\pm$ 0.000 & 0.824 \\
\hline
\textbf{Qwen3-4B}
 & 1  & 0.829 $\pm$ 0.006 & 0.817 $\pm$ 0.009 & 0.802 $\pm$ 0.006 & 0.797 $\pm$ 0.009 & 0.765 $\pm$ 0.007 & 0.815 $\pm$ 0.007 & 0.804 \\
 & 5  & 0.850 $\pm$ 0.004 & 0.842 $\pm$ 0.006 & 0.824 $\pm$ 0.004 & 0.815 $\pm$ 0.004 & 0.792 $\pm$ 0.006 & 0.840 $\pm$ 0.005 & 0.827 \\
 & 10 & 0.853 $\pm$ 0.003 & 0.845 $\pm$ 0.004 & 0.829 $\pm$ 0.005 & 0.819 $\pm$ 0.004 & 0.795 $\pm$ 0.005 & 0.841 $\pm$ 0.004 & 0.830 \\
 & 15 & 0.856 $\pm$ 0.003 & 0.847 $\pm$ 0.002 & 0.829 $\pm$ 0.003 & 0.819 $\pm$ 0.002 & 0.795 $\pm$ 0.002 & 0.843 $\pm$ 0.002 & 0.832 \\
 & 20 & 0.858 $\pm$ 0.002 & 0.845 $\pm$ 0.002 & 0.830 $\pm$ 0.003 & 0.821 $\pm$ 0.003 & 0.793 $\pm$ 0.002 & 0.845 $\pm$ 0.003 & 0.832 \\
 & 25 & 0.858 $\pm$ 0.000 & 0.840 $\pm$ 0.000 & 0.831 $\pm$ 0.000 & 0.818 $\pm$ 0.000 & 0.795 $\pm$ 0.000 & 0.844 $\pm$ 0.000 & 0.831 \\
\hline
\textbf{Gemma-3-12B}
 & 1  & 0.850 $\pm$ 0.005 & 0.845 $\pm$ 0.008 & 0.843 $\pm$ 0.008 & 0.829 $\pm$ 0.006 & 0.793 $\pm$ 0.008 & 0.850 $\pm$ 0.006 & 0.835 \\
 & 5  & 0.866 $\pm$ 0.004 & 0.864 $\pm$ 0.005 & 0.870 $\pm$ 0.004 & 0.851 $\pm$ 0.005 & 0.817 $\pm$ 0.005 & 0.868 $\pm$ 0.004 & 0.856 \\
 & 10 & 0.870 $\pm$ 0.003 & 0.867 $\pm$ 0.004 & 0.873 $\pm$ 0.003 & 0.856 $\pm$ 0.003 & 0.823 $\pm$ 0.003 & 0.872 $\pm$ 0.003 & 0.860 \\
 & 15 & 0.870 $\pm$ 0.001 & 0.868 $\pm$ 0.004 & 0.878 $\pm$ 0.002 & 0.860 $\pm$ 0.003 & 0.825 $\pm$ 0.002 & 0.874 $\pm$ 0.002 & 0.862 \\
 & 20 & 0.870 $\pm$ 0.001 & 0.868 $\pm$ 0.002 & 0.878 $\pm$ 0.002 & 0.859 $\pm$ 0.001 & 0.828 $\pm$ 0.002 & 0.874 $\pm$ 0.002 & 0.863 \\
 & 25 & \textbf{0.871 $\pm$ 0.000} & \textbf{0.868 $\pm$ 0.000} & \textbf{0.879 $\pm$ 0.000} & \textbf{0.860 $\pm$ 0.000} & \textbf{0.826 $\pm$ 0.000} & \textbf{0.880 $\pm$ 0.000} & \textbf{0.864} \\
\hline
\textbf{Gemma-3-4B}
 & 1  & 0.731 $\pm$ 0.008 & 0.723 $\pm$ 0.010 & 0.709 $\pm$ 0.011 & 0.707 $\pm$ 0.008 & 0.650 $\pm$ 0.009 & 0.713 $\pm$ 0.009 & 0.706 \\
 & 5  & 0.767 $\pm$ 0.006 & 0.754 $\pm$ 0.006 & 0.748 $\pm$ 0.007 & 0.743 $\pm$ 0.008 & 0.686 $\pm$ 0.006 & 0.745 $\pm$ 0.006 & 0.740 \\
 & 10 & 0.774 $\pm$ 0.004 & 0.765 $\pm$ 0.004 & 0.757 $\pm$ 0.005 & 0.751 $\pm$ 0.007 & 0.695 $\pm$ 0.005 & 0.753 $\pm$ 0.005 & 0.749 \\
 & 15 & 0.779 $\pm$ 0.003 & 0.769 $\pm$ 0.003 & 0.760 $\pm$ 0.003 & 0.756 $\pm$ 0.004 & 0.699 $\pm$ 0.004 & 0.757 $\pm$ 0.004 & 0.753 \\
 & 20 & 0.781 $\pm$ 0.001 & 0.767 $\pm$ 0.003 & 0.763 $\pm$ 0.002 & 0.758 $\pm$ 0.003 & 0.699 $\pm$ 0.002 & 0.760 $\pm$ 0.002 & 0.755 \\
 & 25 & 0.775 $\pm$ 0.000 & 0.771 $\pm$ 0.000 & 0.765 $\pm$ 0.000 & 0.759 $\pm$ 0.000 & 0.701 $\pm$ 0.000 & 0.762 $\pm$ 0.000 & 0.755 \\
\hline
\end{tabular}
} 
\caption{Aggregated accuracy (mean $\pm$ std) on the \textbf{MMLU} benchmark for each model and agent count under different noise conditions. Only the global best values per column are bolded. Larger models (e.g., Qwen3-14B, Gemma-3-12B) achieve the highest accuracies, while smaller ones (e.g., Qwen3-4B, Gemma-3-4B) show steeper gains with increasing collaboration. Mistral-7B performs lowest across all settings.}
\label{tab:mmlu_accuracy_table}
\end{table*}

\begin{table*}[t]
\centering
\resizebox{\textwidth}{!}{%
\begin{tabular}{l c c c c c c c|c}
\hline
\textbf{Model} & \textbf{Agent Count} & \textbf{Clean} & \textbf{Punct-10} & \textbf{Punct-30} & \textbf{Punct-50} & \textbf{WikiTypo} & \textbf{R\textsuperscript{2}ATA} & \textbf{Avg} \\
\hline
\textbf{Llama-3.1}
 & 1  & $0.233 \pm 0.029$ & $0.215 \pm 0.027$ & $0.211 \pm 0.028$ & $0.203 \pm 0.032$ & $0.185 \pm 0.036$ & $0.237 \pm 0.024$ & $0.214$ \\
 & 5  & $0.370 \pm 0.027$ & $0.330 \pm 0.021$ & $0.311 \pm 0.025$ & $0.308 \pm 0.030$ & $0.287 \pm 0.029$ & $0.348 \pm 0.022$ & $0.326$ \\
 & 10 & $0.466 \pm 0.016$ & $0.416 \pm 0.017$ & $0.367 \pm 0.021$ & $0.386 \pm 0.020$ & $0.348 \pm 0.023$ & $0.427 \pm 0.021$ & $0.402$ \\
 & 15 & $0.506 \pm 0.019$ & $0.435 \pm 0.011$ & $0.393 \pm 0.015$ & $0.407 \pm 0.017$ & $0.370 \pm 0.019$ & $0.457 \pm 0.019$ & $0.428$ \\
 & 20 & $0.508 \pm 0.020$ & $0.453 \pm 0.011$ & $0.410 \pm 0.010$ & $0.410 \pm 0.016$ & $0.385 \pm 0.014$ & $0.473 \pm 0.014$ & $0.440$ \\
 & 25 & $0.540 \pm 0.000$ & $0.480 \pm 0.000$ & $0.440 \pm 0.000$ & $0.410 \pm 0.000$ & $0.370 \pm 0.000$ & $0.490 \pm 0.000$ & $0.455$ \\
\hline
\textbf{Mistral-7B}
 & 1  & $0.068 \pm 0.020$ & $0.057 \pm 0.019$ & $0.058 \pm 0.019$ & $0.049 \pm 0.023$ & $0.042 \pm 0.017$ & $0.055 \pm 0.016$ & $0.055$ \\
 & 5  & $0.100 \pm 0.022$ & $0.078 \pm 0.020$ & $0.084 \pm 0.018$ & $0.069 \pm 0.026$ & $0.066 \pm 0.023$ & $0.073 \pm 0.016$ & $0.078$ \\
 & 10 & $0.148 \pm 0.017$ & $0.099 \pm 0.022$ & $0.125 \pm 0.015$ & $0.101 \pm 0.027$ & $0.089 \pm 0.019$ & $0.106 \pm 0.017$ & $0.111$ \\
 & 15 & $0.156 \pm 0.009$ & $0.123 \pm 0.011$ & $0.143 \pm 0.015$ & $0.112 \pm 0.022$ & $0.105 \pm 0.011$ & $0.115 \pm 0.020$ & $0.126$ \\
 & 20 & $0.168 \pm 0.013$ & $0.120 \pm 0.008$ & $0.152 \pm 0.004$ & $0.103 \pm 0.014$ & $0.123 \pm 0.009$ & $0.122 \pm 0.009$ & $0.131$ \\
 & 25 & $0.180 \pm 0.000$ & $0.130 \pm 0.000$ & $0.140 \pm 0.000$ & $0.100 \pm 0.000$ & $0.120 \pm 0.000$ & $0.130 \pm 0.000$ & $0.133$ \\
\hline
\textbf{Qwen3-14B}
 & 1  & $0.522 \pm 0.032$ & $0.476 \pm 0.027$ & $0.472 \pm 0.029$ & $0.477 \pm 0.031$ & $0.444 \pm 0.025$ & $0.486 \pm 0.021$ & $0.479$ \\
 & 5  & $0.629 \pm 0.026$ & $0.570 \pm 0.021$ & $0.577 \pm 0.023$ & $0.567 \pm 0.021$ & $0.534 \pm 0.018$ & $0.565 \pm 0.018$ & $0.574$ \\
 & 10 & $0.689 \pm 0.016$ & $0.646 \pm 0.017$ & $0.640 \pm 0.014$ & $0.631 \pm 0.018$ & $0.582 \pm 0.011$ & $0.613 \pm 0.020$ & $0.633$ \\
 & 15 & $0.728 \pm 0.011$ & $0.685 \pm 0.028$ & $0.661 \pm 0.012$ & $0.666 \pm 0.009$ & $0.619 \pm 0.011$ & $0.655 \pm 0.013$ & $0.669$ \\
 & 20 & $0.757 \pm 0.011$ & $0.710 \pm 0.013$ & $0.675 \pm 0.016$ & $0.685 \pm 0.013$ & $0.645 \pm 0.005$ & $0.682 \pm 0.012$ & $0.692$ \\
 & 25 & $\mathbf{0.770 \pm 0.000}$ & $\mathbf{0.710 \pm 0.000}$ & $\mathbf{0.690 \pm 0.000}$ & $\mathbf{0.700 \pm 0.000}$ & $\mathbf{0.660 \pm 0.000}$ & $\mathbf{0.700 \pm 0.000}$ & $\mathbf{0.705}$ \\
\hline
\textbf{Qwen3-4B}
 & 1  & $0.524 \pm 0.035$ & $0.481 \pm 0.034$ & $0.484 \pm 0.038$ & $0.489 \pm 0.027$ & $0.410 \pm 0.037$ & $0.488 \pm 0.025$ & $0.479$ \\
 & 5  & $0.649 \pm 0.017$ & $0.607 \pm 0.027$ & $0.602 \pm 0.033$ & $0.592 \pm 0.016$ & $0.535 \pm 0.023$ & $0.610 \pm 0.023$ & $0.599$ \\
 & 10 & $0.688 \pm 0.012$ & $0.657 \pm 0.023$ & $0.656 \pm 0.018$ & $0.645 \pm 0.012$ & $0.597 \pm 0.013$ & $0.657 \pm 0.013$ & $0.650$ \\
 & 15 & $0.714 \pm 0.009$ & $0.680 \pm 0.013$ & $0.676 \pm 0.015$ & $0.658 \pm 0.018$ & $0.625 \pm 0.007$ & $0.698 \pm 0.022$ & $0.675$ \\
 & 20 & $0.722 \pm 0.011$ & $0.702 \pm 0.020$ & $0.688 \pm 0.011$ & $0.692 \pm 0.011$ & $0.638 \pm 0.004$ & $0.715 \pm 0.014$ & $0.693$ \\
 & 25 & $0.730 \pm 0.000$ & $\mathbf{0.740 \pm 0.000}$ & $0.700 \pm 0.000$ & $0.690 \pm 0.000$ & $0.650 \pm 0.000$ & $\mathbf{0.740 \pm 0.000}$ & $0.708$ \\
\hline
\textbf{Gemma-3-12B}
 & 1  & $0.604 \pm 0.031$ & $0.609 \pm 0.024$ & $0.612 \pm 0.028$ & $0.595 \pm 0.027$ & $0.553 \pm 0.033$ & $0.597 \pm 0.033$ & $0.595$ \\
 & 5  & $0.645 \pm 0.024$ & $0.644 \pm 0.017$ & $0.650 \pm 0.025$ & $0.622 \pm 0.020$ & $0.582 \pm 0.022$ & $0.641 \pm 0.030$ & $0.631$ \\
 & 10 & $0.665 \pm 0.017$ & $0.651 \pm 0.019$ & $0.668 \pm 0.016$ & $0.635 \pm 0.011$ & $0.594 \pm 0.010$ & $0.659 \pm 0.014$ & $0.645$ \\
 & 15 & $0.667 \pm 0.014$ & $0.656 \pm 0.014$ & $0.676 \pm 0.010$ & $0.635 \pm 0.015$ & $0.603 \pm 0.009$ & $0.664 \pm 0.019$ & $0.650$ \\
 & 20 & $0.683 \pm 0.016$ & $0.650 \pm 0.010$ & $0.678 \pm 0.007$ & $0.633 \pm 0.016$ & $0.608 \pm 0.007$ & $0.670 \pm 0.016$ & $0.654$ \\
 & 25 & $\mathbf{0.680 \pm 0.000}$ & $0.660 \pm 0.000$ & $\mathbf{0.680 \pm 0.000}$ & $\mathbf{0.650 \pm 0.000}$ & $\mathbf{0.600 \pm 0.000}$ & $0.660 \pm 0.000$ & $\mathbf{0.655}$ \\
\hline
\textbf{Gemma-3-4B}
 & 1  & $0.534 \pm 0.033$ & $0.488 \pm 0.026$ & $0.473 \pm 0.030$ & $0.458 \pm 0.026$ & $0.400 \pm 0.037$ & $0.447 \pm 0.038$ & $0.467$ \\
 & 5  & $0.629 \pm 0.026$ & $0.575 \pm 0.018$ & $0.568 \pm 0.028$ & $0.554 \pm 0.021$ & $0.488 \pm 0.033$ & $0.556 \pm 0.024$ & $0.562$ \\
 & 10 & $0.670 \pm 0.013$ & $0.606 \pm 0.018$ & $0.594 \pm 0.016$ & $0.567 \pm 0.017$ & $0.509 \pm 0.018$ & $0.611 \pm 0.017$ & $0.593$ \\
 & 15 & $0.682 \pm 0.009$ & $0.616 \pm 0.012$ & $0.605 \pm 0.011$ & $0.585 \pm 0.019$ & $0.510 \pm 0.012$ & $0.624 \pm 0.014$ & $0.604$ \\
 & 20 & $0.692 \pm 0.007$ & $0.612 \pm 0.007$ & $0.622 \pm 0.012$ & $0.583 \pm 0.007$ & $0.520 \pm 0.014$ & $0.623 \pm 0.009$ & $0.609$ \\
 & 25 & $0.680 \pm 0.000$ & $0.620 \pm 0.000$ & $0.630 \pm 0.000$ & $0.590 \pm 0.000$ & $0.530 \pm 0.000$ & $0.620 \pm 0.000$ & $0.612$ \\
\hline
\end{tabular}
} 
\caption{Aggregated accuracy (mean $\pm$ std) on the \textbf{Math} benchmark for each model and agent count under different noise conditions. Only the single highest value per column is bolded. Overall, Qwen3-14B achieves the top accuracy in most categories, followed closely by Gemma-3-12B.}
\label{tab:math_accuracy_table}
\end{table*}

\begin{table*}[t]
\centering
\resizebox{\textwidth}{!}{%
\begin{tabular}{l c c c c c c c|c}
\hline
\textbf{Model} & \textbf{Agent Count} & \textbf{Clean} & \textbf{Punct-10} & \textbf{Punct-30} & \textbf{Punct-50} & \textbf{WikiTypo} & \textbf{R\textsuperscript{2}ATA} & \textbf{Avg} \\
\hline
\textbf{Llama-3.1}
 & 1  & $0.632 \pm 0.011$ & $0.637 \pm 0.011$ & $0.624 \pm 0.010$ & $0.614 \pm 0.010$ & $0.466 \pm 0.012$ & $0.608 \pm 0.013$ & $0.597$ \\
 & 5  & $0.799 \pm 0.018$ & $0.788 \pm 0.013$ & $0.763 \pm 0.013$ & $0.750 \pm 0.014$ & $0.588 \pm 0.016$ & $0.766 \pm 0.020$ & $0.742$ \\
 & 10 & $0.886 \pm 0.004$ & $0.869 \pm 0.005$ & $0.846 \pm 0.008$ & $0.834 \pm 0.005$ & $0.668 \pm 0.007$ & $0.843 \pm 0.007$ & $0.825$ \\
 & 15 & $0.905 \pm 0.003$ & $0.886 \pm 0.002$ & $0.866 \pm 0.005$ & $0.856 \pm 0.004$ & $0.695 \pm 0.005$ & $0.865 \pm 0.004$ & $0.846$ \\
 & 20 & $0.911 \pm 0.001$ & $0.891 \pm 0.002$ & $0.874 \pm 0.002$ & $0.867 \pm 0.002$ & $0.708 \pm 0.003$ & $0.877 \pm 0.003$ & $0.855$ \\
 & 25 & $0.913 \pm 0.000$ & $0.901 \pm 0.000$ & $0.877 \pm 0.000$ & $0.873 \pm 0.000$ & $0.713 \pm 0.000$ & $0.877 \pm 0.000$ & $0.859$ \\
\hline
\textbf{Mistral-7B}
 & 1  & $0.428 \pm 0.010$ & $0.397 \pm 0.009$ & $0.368 \pm 0.011$ & $0.338 \pm 0.015$ & $0.264 \pm 0.010$ & $0.385 \pm 0.013$ & $0.363$ \\
 & 5  & $0.553 \pm 0.019$ & $0.512 \pm 0.021$ & $0.484 \pm 0.024$ & $0.433 \pm 0.018$ & $0.341 \pm 0.013$ & $0.490 \pm 0.019$ & $0.469$ \\
 & 10 & $0.659 \pm 0.009$ & $0.614 \pm 0.008$ & $0.582 \pm 0.013$ & $0.538 \pm 0.005$ & $0.420 \pm 0.007$ & $0.594 \pm 0.008$ & $0.568$ \\
 & 15 & $0.692 \pm 0.004$ & $0.657 \pm 0.003$ & $0.621 \pm 0.008$ & $0.581 \pm 0.005$ & $0.454 \pm 0.004$ & $0.632 \pm 0.006$ & $0.606$ \\
 & 20 & $0.718 \pm 0.002$ & $0.678 \pm 0.007$ & $0.641 \pm 0.003$ & $0.607 \pm 0.003$ & $0.476 \pm 0.006$ & $0.652 \pm 0.004$ & $0.629$ \\
 & 25 & $0.732 \pm 0.000$ & $0.690 \pm 0.000$ & $0.656 \pm 0.000$ & $0.618 \pm 0.000$ & $0.480 \pm 0.000$ & $0.657 \pm 0.000$ & $0.639$ \\
\hline
\textbf{Qwen3-14B}
 & 1  & $0.875 \pm 0.005$ & $0.868 \pm 0.006$ & $0.861 \pm 0.006$ & $0.862 \pm 0.005$ & $0.750 \pm 0.009$ & $0.863 \pm 0.006$ & $0.846$ \\
 & 5  & $0.921 \pm 0.003$ & $0.917 \pm 0.003$ & $0.915 \pm 0.004$ & $0.918 \pm 0.004$ & $0.808 \pm 0.006$ & $0.906 \pm 0.004$ & $0.898$ \\
 & 10 & $0.931 \pm 0.003$ & $0.928 \pm 0.003$ & $0.924 \pm 0.002$ & $0.929 \pm 0.002$ & $0.824 \pm 0.003$ & $0.918 \pm 0.002$ & $0.909$ \\
 & 15 & $0.933 \pm 0.002$ & $0.931 \pm 0.001$ & $0.926 \pm 0.002$ & $0.932 \pm 0.002$ & $0.830 \pm 0.002$ & $0.920 \pm 0.001$ & $0.912$ \\
 & 20 & $0.934 \pm 0.001$ & $0.932 \pm 0.001$ & $0.927 \pm 0.001$ & $0.933 \pm 0.002$ & $0.833 \pm 0.003$ & $0.921 \pm 0.002$ & $0.913$ \\
 & 25 & $0.934 \pm 0.000$ & $\mathbf{0.934 \pm 0.000}$ & $0.929 \pm 0.000$ & $\mathbf{0.934 \pm 0.000}$ & $\mathbf{0.837 \pm 0.000}$ & $0.924 \pm 0.000$ & $\mathbf{0.915}$ \\
\hline
\textbf{Qwen3-4B}
 & 1  & $0.819 \pm 0.007$ & $0.790 \pm 0.009$ & $0.777 \pm 0.009$ & $0.788 \pm 0.008$ & $0.573 \pm 0.010$ & $0.761 \pm 0.007$ & $0.751$ \\
 & 5  & $0.898 \pm 0.006$ & $0.880 \pm 0.006$ & $0.869 \pm 0.007$ & $0.872 \pm 0.007$ & $0.670 \pm 0.010$ & $0.846 \pm 0.010$ & $0.839$ \\
 & 10 & $0.916 \pm 0.002$ & $0.905 \pm 0.005$ & $0.890 \pm 0.004$ & $0.899 \pm 0.004$ & $0.702 \pm 0.004$ & $0.868 \pm 0.006$ & $0.863$ \\
 & 15 & $0.922 \pm 0.002$ & $0.913 \pm 0.003$ & $0.895 \pm 0.003$ & $0.906 \pm 0.002$ & $0.712 \pm 0.005$ & $0.873 \pm 0.005$ & $0.870$ \\
 & 20 & $0.925 \pm 0.002$ & $0.914 \pm 0.001$ & $0.900 \pm 0.002$ & $0.910 \pm 0.002$ & $0.717 \pm 0.003$ & $0.878 \pm 0.002$ & $0.874$ \\
 & 25 & $0.930 \pm 0.000$ & $0.916 \pm 0.000$ & $0.899 \pm 0.000$ & $0.914 \pm 0.000$ & $0.722 \pm 0.000$ & $0.882 \pm 0.000$ & $0.877$ \\
\hline
\textbf{Gemma-3-12B}
 & 1  & $0.911 \pm 0.004$ & $0.907 \pm 0.005$ & $0.900 \pm 0.005$ & $0.897 \pm 0.005$ & $0.740 \pm 0.007$ & $0.898 \pm 0.004$ & $0.875$ \\
 & 5  & $0.935 \pm 0.002$ & $0.930 \pm 0.002$ & $0.921 \pm 0.004$ & $0.918 \pm 0.004$ & $0.783 \pm 0.006$ & $0.919 \pm 0.004$ & $0.901$ \\
 & 10 & $0.938 \pm 0.002$ & $0.933 \pm 0.002$ & $0.924 \pm 0.002$ & $0.921 \pm 0.002$ & $0.796 \pm 0.004$ & $0.923 \pm 0.002$ & $0.906$ \\
 & 15 & $0.941 \pm 0.002$ & $0.933 \pm 0.001$ & $0.925 \pm 0.003$ & $0.923 \pm 0.002$ & $0.801 \pm 0.004$ & $0.924 \pm 0.002$ & $0.908$ \\
 & 20 & $0.941 \pm 0.001$ & $0.935 \pm 0.001$ & $0.927 \pm 0.001$ & $0.922 \pm 0.001$ & $0.802 \pm 0.001$ & $0.924 \pm 0.001$ & $0.909$ \\
 & 25 & $\mathbf{0.942 \pm 0.000}$ & $0.933 \pm 0.000$ & $\mathbf{0.928 \pm 0.000}$ & $0.923 \pm 0.000$ & $0.802 \pm 0.000$ & $\mathbf{0.924 \pm 0.000}$ & $\mathbf{0.909}$ \\
\hline
\textbf{Gemma-3-4B}
 & 1  & $0.849 \pm 0.006$ & $0.840 \pm 0.007$ & $0.817 \pm 0.009$ & $0.803 \pm 0.006$ & $0.582 \pm 0.007$ & $0.823 \pm 0.008$ & $0.786$ \\
 & 5  & $0.898 \pm 0.004$ & $0.889 \pm 0.005$ & $0.874 \pm 0.005$ & $0.857 \pm 0.006$ & $0.651 \pm 0.005$ & $0.873 \pm 0.004$ & $0.840$ \\
 & 10 & $0.906 \pm 0.002$ & $0.896 \pm 0.003$ & $0.882 \pm 0.003$ & $0.869 \pm 0.004$ & $0.672 \pm 0.006$ & $0.883 \pm 0.003$ & $0.852$ \\
 & 15 & $0.908 \pm 0.001$ & $0.897 \pm 0.001$ & $0.887 \pm 0.002$ & $0.871 \pm 0.003$ & $0.676 \pm 0.003$ & $0.886 \pm 0.003$ & $0.854$ \\
 & 20 & $0.909 \pm 0.001$ & $0.899 \pm 0.001$ & $0.888 \pm 0.001$ & $0.873 \pm 0.002$ & $0.676 \pm 0.002$ & $0.888 \pm 0.001$ & $0.855$ \\
 & 25 & $0.909 \pm 0.000$ & $0.898 \pm 0.000$ & $0.888 \pm 0.000$ & $0.876 \pm 0.000$ & $0.674 \pm 0.000$ & $0.892 \pm 0.000$ & $0.856$ \\
\hline
\end{tabular}
} 
\caption{Aggregated accuracy (mean $\pm$ std) on the \textbf{GSM8K} benchmark for each model and agent count under different noise conditions. Only the single best value per column is bolded. Gemma-3-12B achieves the highest performance overall, particularly under noisy settings.}
\label{tab:gsm_accuracy_table}
\end{table*}

\begin{table*}[t]
\centering
\resizebox{\textwidth}{!}{%
\begin{tabular}{l c c c c c c c|c}
\hline
\textbf{Model} & \textbf{Agent Count} & \textbf{Clean} & \textbf{Punct-10} & \textbf{Punct-30} & \textbf{Punct-50} & \textbf{WikiTypo} & \textbf{R\textsuperscript{2}ATA} & \textbf{Avg} \\
\hline
\textbf{Llama-3.1}
 & 1  & $0.747 \pm 0.029$ & $0.787 \pm 0.034$ & $0.758 \pm 0.024$ & $0.753 \pm 0.027$ & $0.737 \pm 0.035$ & $0.812 \pm 0.018$ & $0.766$ \\
 & 5  & $0.959 \pm 0.019$ & $0.957 \pm 0.015$ & $0.917 \pm 0.013$ & $0.893 \pm 0.018$ & $0.931 \pm 0.048$ & $0.968 \pm 0.012$ & $0.937$ \\
 & 10 & $0.995 \pm 0.003$ & $0.986 \pm 0.003$ & $0.949 \pm 0.007$ & $0.926 \pm 0.010$ & $0.984 \pm 0.007$ & $0.987 \pm 0.008$ & $0.971$ \\
 & 15 & $0.997 \pm 0.003$ & $0.987 \pm 0.002$ & $0.954 \pm 0.004$ & $0.933 \pm 0.004$ & $0.993 \pm 0.003$ & $0.994 \pm 0.004$ & $0.977$ \\
 & 20 & $0.996 \pm 0.003$ & $0.985 \pm 0.003$ & $0.955 \pm 0.004$ & $0.937 \pm 0.003$ & $0.994 \pm 0.000$ & $0.994 \pm 0.000$ & $0.977$ \\
 & 25 & $1.000 \pm 0.000$ & $0.989 \pm 0.000$ & $0.961 \pm 0.000$ & $0.933 \pm 0.000$ & $1.000 \pm 0.000$ & $0.989 \pm 0.000$ & $0.979$ \\
\hline
\textbf{Mistral-7B}
 & 1  & $0.709 \pm 0.026$ & $0.712 \pm 0.029$ & $0.657 \pm 0.030$ & $0.620 \pm 0.034$ & $0.711 \pm 0.035$ & $0.696 \pm 0.030$ & $0.684$ \\
 & 5  & $0.913 \pm 0.012$ & $0.911 \pm 0.013$ & $0.853 \pm 0.020$ & $0.830 \pm 0.020$ & $0.914 \pm 0.022$ & $0.879 \pm 0.019$ & $0.883$ \\
 & 10 & $0.962 \pm 0.008$ & $0.959 \pm 0.015$ & $0.919 \pm 0.011$ & $0.902 \pm 0.012$ & $0.976 \pm 0.012$ & $0.930 \pm 0.007$ & $0.942$ \\
 & 15 & $0.966 \pm 0.011$ & $0.975 \pm 0.010$ & $0.939 \pm 0.007$ & $0.926 \pm 0.010$ & $0.988 \pm 0.006$ & $0.943 \pm 0.008$ & $0.956$ \\
 & 20 & $0.976 \pm 0.007$ & $0.981 \pm 0.003$ & $0.947 \pm 0.007$ & $0.939 \pm 0.003$ & $0.988 \pm 0.004$ & $0.951 \pm 0.005$ & $0.964$ \\
 & 25 & $0.989 \pm 0.000$ & $0.983 \pm 0.000$ & $0.950 \pm 0.000$ & $0.939 \pm 0.000$ & $1.000 \pm 0.000$ & $0.956 \pm 0.000$ & $0.969$ \\
\hline
\textbf{Qwen3-14B}
 & 1  & $0.982 \pm 0.007$ & $0.969 \pm 0.011$ & $0.974 \pm 0.011$ & $0.965 \pm 0.011$ & $0.978 \pm 0.010$ & $0.986 \pm 0.005$ & $0.976$ \\
 & 5  & $0.996 \pm 0.002$ & $0.983 \pm 0.004$ & $0.990 \pm 0.005$ & $0.989 \pm 0.007$ & $0.997 \pm 0.003$ & $0.990 \pm 0.003$ & $0.991$ \\
 & 10 & $0.996 \pm 0.002$ & $0.984 \pm 0.002$ & $0.990 \pm 0.005$ & $0.994 \pm 0.004$ & $0.998 \pm 0.003$ & $0.991 \pm 0.003$ & $0.992$ \\
 & 15 & $0.995 \pm 0.002$ & $0.983 \pm 0.000$ & $0.992 \pm 0.004$ & $0.993 \pm 0.003$ & $0.999 \pm 0.002$ & $0.991 \pm 0.003$ & $0.992$ \\
 & 20 & $0.994 \pm 0.000$ & $0.983 \pm 0.000$ & $0.994 \pm 0.002$ & $0.994 \pm 0.000$ & $0.997 \pm 0.003$ & $0.994 \pm 0.002$ & $0.993$ \\
 & 25 & $0.994 \pm 0.000$ & $0.983 \pm 0.000$ & $0.994 \pm 0.000$ & $0.994 \pm 0.000$ & $\mathbf{1.000 \pm 0.000}$ & $0.994 \pm 0.000$ & $0.994$ \\
\hline
\textbf{Qwen3-4B}
 & 1  & $0.972 \pm 0.009$ & $0.940 \pm 0.019$ & $0.902 \pm 0.021$ & $0.916 \pm 0.017$ & $0.968 \pm 0.014$ & $0.931 \pm 0.017$ & $0.938$ \\
 & 5  & $0.996 \pm 0.003$ & $0.997 \pm 0.004$ & $0.967 \pm 0.007$ & $0.972 \pm 0.006$ & $0.998 \pm 0.003$ & $0.984 \pm 0.005$ & $0.986$ \\
 & 10 & $0.997 \pm 0.003$ & $\mathbf{1.000 \pm 0.001}$ & $0.970 \pm 0.006$ & $0.977 \pm 0.006$ & $\mathbf{1.000 \pm 0.001}$ & $0.984 \pm 0.002$ & $0.988$ \\
 & 15 & $0.999 \pm 0.002$ & $1.000 \pm 0.000$ & $0.974 \pm 0.003$ & $0.973 \pm 0.002$ & $1.000 \pm 0.000$ & $0.984 \pm 0.002$ & $0.988$ \\
 & 20 & $\mathbf{1.000 \pm 0.000}$ & $1.000 \pm 0.000$ & $0.974 \pm 0.003$ & $0.973 \pm 0.002$ & $1.000 \pm 0.000$ & $0.983 \pm 0.000$ & $0.988$ \\
 & 25 & $1.000 \pm 0.000$ & $1.000 \pm 0.000$ & $0.978 \pm 0.000$ & $0.972 \pm 0.000$ & $1.000 \pm 0.000$ & $0.983 \pm 0.000$ & $0.989$ \\
\hline
\textbf{Gemma-3-12B}
 & 1  & $\mathbf{0.998 \pm 0.003}$ & $0.979 \pm 0.009$ & $0.967 \pm 0.009$ & $0.950 \pm 0.012$ & $0.999 \pm 0.002$ & $0.991 \pm 0.004$ & $0.981$ \\
 & 5  & $\mathbf{1.000 \pm 0.000}$ & $0.986 \pm 0.006$ & $0.978 \pm 0.004$ & $0.960 \pm 0.009$ & $\mathbf{1.000 \pm 0.000}$ & $\mathbf{0.993 \pm 0.003}$ & $0.986$ \\
 & 10 & $\mathbf{1.000 \pm 0.000}$ & $0.990 \pm 0.003$ & $0.980 \pm 0.003$ & $0.965 \pm 0.008$ & $\mathbf{1.000 \pm 0.000}$ & $\mathbf{0.993 \pm 0.002}$ & $0.988$ \\
 & 15 & $\mathbf{1.000 \pm 0.000}$ & $0.991 \pm 0.004$ & $\mathbf{0.983 \pm 0.002}$ & $0.970 \pm 0.005$ & $\mathbf{1.000 \pm 0.000}$ & $\mathbf{0.994 \pm 0.000}$ & $\mathbf{0.990}$ \\
 & 20 & $\mathbf{1.000 \pm 0.000}$ & $0.990 \pm 0.004$ & $0.981 \pm 0.003$ & $0.971 \pm 0.004$ & $\mathbf{1.000 \pm 0.000}$ & $\mathbf{0.994 \pm 0.000}$ & $\mathbf{0.990}$ \\
 & 25 & $\mathbf{1.000 \pm 0.000}$ & $0.994 \pm 0.000$ & $0.983 \pm 0.000$ & $0.967 \pm 0.000$ & $\mathbf{1.000 \pm 0.000}$ & $\mathbf{0.994 \pm 0.000}$ & $\mathbf{0.990}$ \\
\hline
\textbf{Gemma-3-4B}
 & 1  & $0.969 \pm 0.010$ & $0.929 \pm 0.013$ & $0.934 \pm 0.012$ & $0.930 \pm 0.012$ & $0.970 \pm 0.011$ & $0.942 \pm 0.011$ & $0.946$ \\
 & 5  & $0.993 \pm 0.005$ & $0.951 \pm 0.008$ & $0.960 \pm 0.007$ & $0.953 \pm 0.005$ & $0.995 \pm 0.004$ & $0.954 \pm 0.006$ & $0.968$ \\
 & 10 & $0.997 \pm 0.003$ & $0.952 \pm 0.004$ & $0.966 \pm 0.007$ & $0.955 \pm 0.001$ & $0.995 \pm 0.003$ & $0.952 \pm 0.006$ & $0.969$ \\
 & 15 & $0.998 \pm 0.002$ & $0.952 \pm 0.002$ & $0.967 \pm 0.008$ & $0.956 \pm 0.002$ & $0.997 \pm 0.003$ & $0.954 \pm 0.004$ & $0.971$ \\
 & 20 & $0.999 \pm 0.002$ & $0.950 \pm 0.000$ & $0.966 \pm 0.004$ & $0.956 \pm 0.000$ & $0.999 \pm 0.002$ & $0.955 \pm 0.004$ & $0.971$ \\
 & 25 & $1.000 \pm 0.000$ & $0.950 \pm 0.000$ & $0.972 \pm 0.000$ & $0.956 \pm 0.000$ & $1.000 \pm 0.000$ & $0.956 \pm 0.000$ & $0.972$ \\
\hline
\end{tabular}
} 
\caption{Aggregated accuracy (mean $\pm$ std) on the \textbf{MultiArith} benchmark for each model and agent count under different noise conditions. Only the single highest value per column is bolded. Gemma-3-12B consistently reaches or matches the top accuracy across all conditions.}
\label{tab:multiarith_accuracy_table}
\end{table*}

\end{document}